\documentclass[journal]{IEEEtran}

\usepackage{times}
\usepackage{graphicx}
\usepackage{epsfig}
\usepackage{float}
\usepackage{amsmath}
\usepackage{rotating}
\usepackage{amssymb}
\usepackage{array}
\usepackage{multirow}
\usepackage{caption}
\usepackage{subcaption}
\newcolumntype{L}[1]{>{\raggedright\let\newline\\\arraybackslash\hspace{0pt}}m{#1}}
\newcolumntype{C}[1]{>{\centering\let\newline\\\arraybackslash\hspace{0pt}}m{#1}}
\newcolumntype{R}[1]{>{\raggedleft\let\newline\\\arraybackslash\hspace{0pt}}m{#1}}

\usepackage[bookmarks=false]{hyperref}
\DeclareMathOperator*{\argmin}{arg\,min}
\DeclareMathOperator{\Tr}{Tr}

\begin{document}

\title{Supervised Mixed Norm Autoencoder for Kinship Verification in Unconstrained Videos}

\author{Naman Kohli,~\IEEEmembership{Student~Member,~IEEE,}
       Daksha Yadav,~\IEEEmembership{Student~Member,~IEEE,}
        Mayank Vatsa, ~\IEEEmembership{Senior~Member,~IEEE,}
        Richa Singh,~\IEEEmembership{Senior~Member,~IEEE,}
		and Afzel Noore,~\IEEEmembership{Senior~Member,~IEEE.}
\thanks{N. Kohli and D. Yadav are with Lane Department of Computer Science
and Electrical Engineering, West Virginia University, Morgantown, WV, USA 26506 (e-mail: nakohli@mix.wvu.edu; dayadav@mix.wvu.edu)}
\thanks{M. Vatsa and R. Singh are with IIIT-Delhi, New Delhi, India 110028 (e-mail: mayank@iiitd.ac.in;
rsingh@iiitd.ac.in)}
\thanks{A. Noore is with Texas A\&M University-Kingsville, Kingsville, TX, USA 78363 (e-mail: afzel.noore@tamuk.edu)}
\thanks{Manuscript received March, 2017}}

\maketitle

\begin{abstract}
Identifying kinship relations has garnered interest due to several applications such as organizing and tagging the enormous amount of videos being uploaded on the Internet. Existing research in kinship verification primarily focuses on kinship prediction with image pairs. In this research, we propose a new deep learning framework for kinship verification in unconstrained videos using a novel Supervised Mixed Norm regularization Autoencoder (SMNAE). This new autoencoder formulation introduces class-specific sparsity in the weight matrix. The proposed three-stage SMNAE based kinship verification framework utilizes the learned spatio-temporal representation in the video frames for verifying kinship in a pair of videos. A new kinship video (KIVI) database of more than 500 individuals with variations due to illumination, pose, occlusion, ethnicity, and expression is collected for this research. It comprises a total of 355 true kin video pairs with over 250,000 still frames. The effectiveness of the proposed framework is demonstrated on the KIVI database and six existing kinship databases. On the KIVI database, SMNAE yields video-based kinship verification accuracy of 83.18\% which is at least $\textbf{3.2\%}$ better than existing algorithms. The algorithm is also evaluated on six publicly available kinship databases and compared with best reported results. It is observed that the proposed SMNAE consistently yields best results on all the databases.
\end{abstract}

\section{Introduction}

\begingroup
\renewcommand*{\arraystretch}{1}
\begin{table*}[!ht]
\centering
\caption{Kinship verification algorithms and databases published in the literature.}
\label{tab:review_table}
\begin{tabular}{|C{1cm}|L{3.8cm}|L{6cm}|L{5.45cm}|}
\hline
\textbf{Input} & \textbf{Authors (Year)}  & \textbf{Kinship Verification Algorithm} & \textbf{Database}  \\ \hline

\multirow{38}{*}{Image} & Fang et al. \cite{Fang2010} (2010) & Gabor-based gradient orientation pyramid & CornellKin  \\\cline{2-4} 
 
 & Siyu et al. \cite{Siyu2011} (2011) & Transfer learning & UB KinFace  \\\cline{2-4} 

& Zhou et al. \cite{Zhou2011} (2011) & Spatial pyramid based learning & Private database \\\cline{2-4}

 & Shao et al. \cite{Ming_CVPR11_Genealogical} (2011) & Gabor filters with metric learning & UB KinFace  \\\cline{2-4}

& Zhou et al. \cite{Zhou2012} (2012) & Gabor-based gradient orientation pyramid & Private database  \\\cline{2-4} 

& Xia et al. \cite{xia2012understanding} (2012) & Transfer subspace learning based algorithm & UB KinFace Ver2.0 and FamilyFace  \\\cline{2-4} 

& Kohli et al. \cite{Kohli2012} (2012) & Self-similarity representation of Weber faces & UB KinFace and IIITD Kinship \\\cline{2-4}

& Fang et al. \cite{fang2013kinship} (2013) & Reconstruction using parts from a set of families & Family101 \\\cline{2-4}

& Lu et al. \cite{lu} (2014) & Multi-view NRML  & KinFaceW-I and KinFaceW-II  \\\cline{2-4}

& Hu et al. \cite{hu2014large} (2014) & Large margin multi metric learning  & KinFaceW-I and KinFaceW-II  \\\cline{2-4}

& Yan et al.\cite{Yan} (2014) & Discriminative multimetric learning & KinFaceW-I, KinFaceW-II, CornellKin, and UB KinFace \\ \cline{2-4}          

& Dehghan et al. \cite{dehghanOVS14} (2014) & Discrimination via gated autoencoders & KinFaceW-I and KinFaceW-II \\ \cline{2-4}

& Guo et al. \cite{guo2014graph} (2014) & Graph-based approach &  Sibling-Face and Group-Face \\\cline{2-4}

& Yan et al. \cite{Yan2014} (2015) & Prototype discriminative feature learning &  KinFaceW-I, KinFaceW-II, CornellKin, and UB KinFace \\\cline{2-4}

& Liu et al. \cite{fisher_kinship} (2015) & Inheritable Fisher vector feature based kinship & KinFaceW-I and KinFaceW-II \\\cline{2-4}

& Alirezazadeh et al. \cite{genetic_kinship} (2015) & Genetic algorithm for feature selection & KinFaceW-I and KinFaceW-II \\ \cline{2-4}

& Qin et al. \cite{qin} (2015) & Relative symmetric bilinear model and spatially
voted feature selection method & TSKinFace, Family101, KinFaceW-I, and KinFaceW-II \\\cline{2-4}

& Zhou et al. \cite{Zhou2015} (2016)  & Ensemble similarity learning & KinFaceW-I and KinFaceW-II\\\cline{2-4}

& Robinson et al. \cite{robinson2016families} (2016) &  Fine-tuning VGG network & Families in the Wild\\\cline{2-4}

& Xu and Shang \cite{xuandshang} (2016) & Joint learning of multiple bilinear similarity models & KinFaceW-I and KinFaceW-II \\\cline{2-4}

& Wu et al. \cite{wuusefulness} (2016) & Utilized color-texture features & TSKinFace, KinFaceW-I, and KinFaceW-II \\\cline{2-4}

& Yan \cite{Yan2016} (2016) & Neighborhood repulsed correlation metric learning & TSKinFace, KinFaceW-I, and KinFaceW-II \\\cline{2-4}

& Lopez et al. \cite{comments2015} (2016) & Chromaticity and color features &  KinFaceW-I and KinFaceW-II \\\cline{2-4}

& Xu and Shang \cite{xuandshang} (2016) & Used structured similarity fusion & KinFaceW-I and KinFaceW-II \\\cline{2-4}


& Li et al. \cite{LiFWXH16} (2016) & Siamese convolutional neural net & KinFaceW-I and KinFaceW-II \\\cline{2-4}

& Wang et al. \cite{wang2017kinship} (2017) & Denoising auto-encoder based  metric learning &   Families In the Wild (FIW)  \\ \cline{2-4} 

& Lu et al. \cite{lu2017discriminative} (2017) & Discriminative deep multi-metric learning &  KinFaceW-I and KinFaceW-II \\ \cline{2-4} 

& Liu et al. \cite{liu2017} (2017) & Status-aware projection learning &  KinFaceW-I and KinFaceW-II \\ \cline{2-4} 

& Kohli et al. \cite{kohli16} (2017) & Kinship verification via representation learning &  WVU, CornellKin, UB KinFace, KinFaceW-I, and KinFaceW-II  \\ \cline{2-4} 

& Mahpod et al. \cite{mahpod2018} (2018) & Multi-view hybrid distance learning &  CornellKin, KinFaceW-I and KinFaceW-II \\ \cline{2-4} \hline \hline

\multirow{4}{*}{Video} & Dibeklioglu et al. \cite{uvanemo} (2013) & Spatio-temporal features utilizing facial dynamics &  UvA-NEMO Smile \\ \cline{2-4}
&  Dibeklioglu \cite{dibeklioglu2017visual} (2017)  & Visual transformation aided contrastive learning  & UvA-NEMO, KinFaceW-I, and KinFaceW-II \\ \cline{2-4}
&  \textbf{Proposed} (2018)  & Framework utilizing Supervised Mixed Norm Autoencoder & KIVI, UvA-NEMO, WVU, KinFaceW-I, KinFaceW-II, CornellKin, and UB KinFace \\ \hline
\end{tabular} 
\end{table*}
\endgroup

\begin{figure}
\centering
 \includegraphics[width=0.8\linewidth]{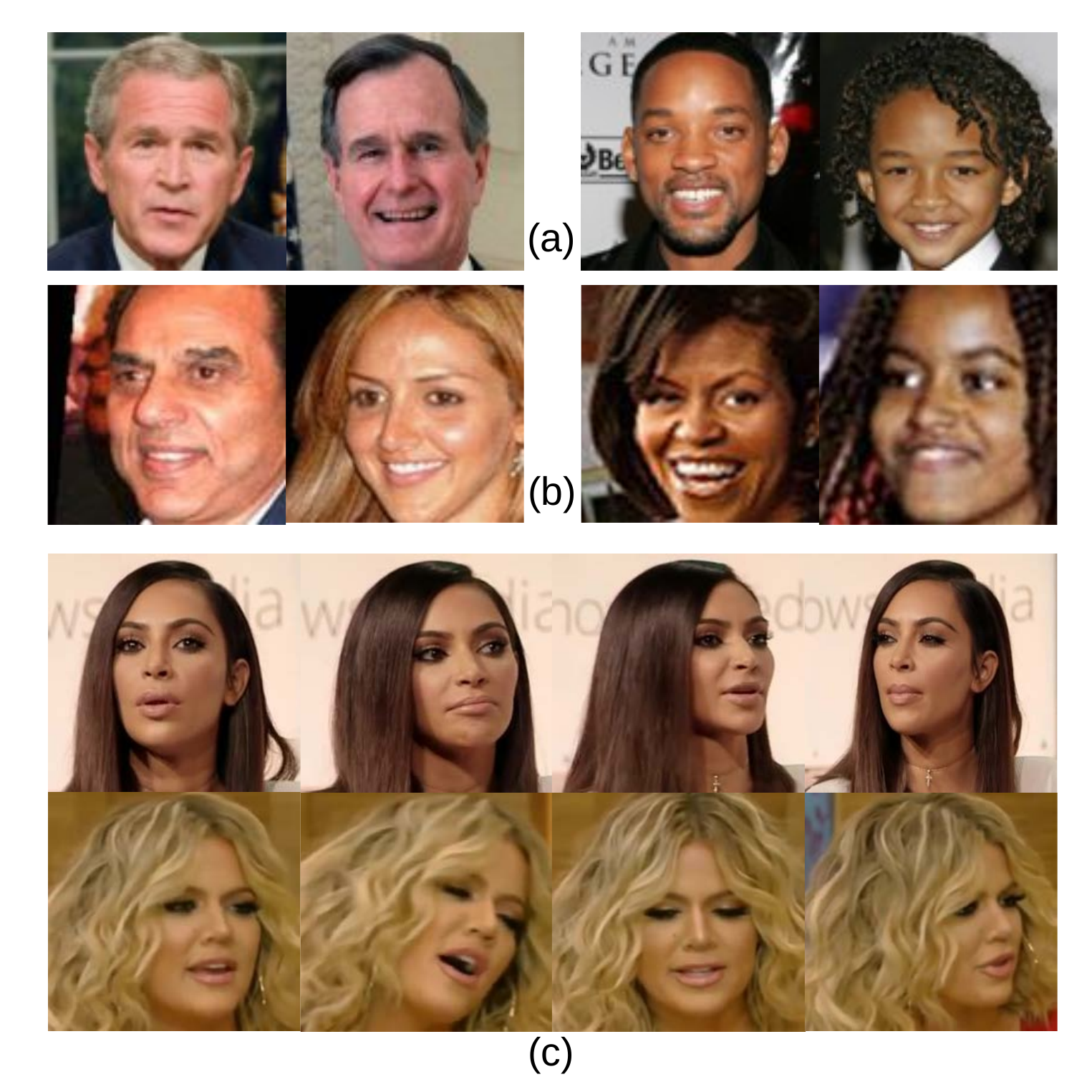}

 
 
  \caption{Progression of research in automatic kinship verification: (a) Cornell KinFace \cite{Fang2010} database (2010), (b) Family 101 database \cite{fang2013kinship} (2013), and (c) proposed KIVI database (2018).} 
  
  \vspace{-2mm}
 \label{fig:motivation}
\end{figure}

Kinship verification using facial images is the process of analyzing facial features to determine if two individuals are biologically related. Recently, kinship verification has received increasing attention from computer vision research community due to its applications ranging from identifying family relationships to indexing images. However, automatic kinship verification in videos is a relatively unexplored research area and can be highly valuable in diverse settings such as security, surveillance, and immigration control. For instance, during the investigation of surveillance footage in the Boston Marathon bombing, two male suspects were determined as the bombers. Later, it was established that the two males were brothers which led to their identification. An automatic kinship verification system that determines kin in a video could have expedited this investigation. Another application of kinship verification using videos is for border control where surveillance videos can be applied to validate the relation between an adult and the child, thereby preventing illegal child trafficking. Additionally, video-based kinship verification can validate or disprove kinship claims of refugees and asylum seekers. Currently, as part of its reunification program, the U.S. State Department conducts DNA testing to allow people who have relatives in the U.S. to enter as refugees \cite{dna}. Rapid-DNA is being used for this purpose; however, an automatic kinship verification algorithm can produce cost-effective results in real-time. In these instances, automatic kinship verification can be beneficial in screening possible fraudulent cases in a non-intrusive manner. 

Kinship information can also be used for managing multimedia on social media websites such as Facebook and Youtube. According to Wagner \cite{stat_Youtube}, in 2016, 500 million people watched Facebook videos every day. In many cases, family members have different Youtube channels where they upload daily videos. Kinship information can be applied for automatically tagging such videos and identifying the kin present in them. Kinship context in videos can also be used for automatic indexing and organization of videos, making them easily searchable.

\subsection{Related Work}
Table \ref{tab:review_table} and Fig. \ref{fig:motivation} showcase the research progression in the area of kinship verification. The problem of kinship verification using facial images was first proposed by Fang et al. \cite{Fang2010}. Since its origin, different frameworks and techniques have been applied for solving this problem which can be classified into two categories: approaches using handcrafted features and approaches based on learning features. Various handcrafted features have been incorporated to model kinship features and similarities in faces. Fang et al. \cite{Fang2010} employed low-level features such as eye color, skin color, and eye-to-nose distance for kinship verification. Zhou et al. \cite{Zhou2011} proposed a new spatial pyramid learning-based descriptor to encode kinship facial features. Zhou et al. \cite{Zhou2012} introduced Gabor-based gradient orientation pyramid representation method to learn kinship information in uncontrolled conditions. Kohli et al. \cite{Kohli2012} computed self-similarity descriptor to propose self-similarity representation of Weber face algorithm to classify a given pair of images as kin or non-kin. Liu et al. \cite{fisher_kinship} computed inheritable Fisher vector feature by applying the inheritable transformation on the Fisher vector for each image and performed kinship verification. Wu et al. \cite{wuusefulness} demonstrated the efficacy of joint color-texture features from the luminance and chrominance channels of color images in kinship verification.

Researchers have also proposed methods for learning kinship cues based on transfer learning, dictionary learning, metric learning, and deep learning for addressing this challenging problem. Shao et al. \cite{Ming_CVPR11_Genealogical} learned a subspace where the similarity between child-old parent and child-young parent pairs was maximized to separate them in the subspace. Siyu et al. \cite{Siyu2011} proposed a transfer subspace learning approach using the young parent set as the intermediate domain whose distribution is close to child and old parent distributions. Fang et al. \cite{fang2013kinship} utilized the knowledge that an offspring shares genetic material with parents and reconstructed the probe face from a sparse set of samples among the candidate families. Dehgan et al. \cite{dehghanOVS14} proposed a framework for kinship verification which combined various features and metrics computed via gated autoencoders with a discriminative neural network layer. Kohli et al. \cite{kohli16} introduced a novel approach for feature representation which encoded kinship information present in images using filtered contractive deep belief network. 

\begingroup
\renewcommand*{\arraystretch}{1.1}
\begin{table}[t]
\centering
\caption{Summary of kinship databases in the literature.}
\label{tab:kindb}
\begin{tabular}{|C{0.72cm}|L{2cm}|C{1.3cm}|C{1.3cm}|C{1.35cm}|}
\hline
                   & \textbf{Dataset}    & \textbf{No. of Subjects} & \textbf{No. of Samples} & \textbf{Relations} \\ \hline

\multirow{8}{*}{Image} & CornellKin \cite{Fang2010}& 300& 300 & 4 \\ \cline{2-5} 
                        & UB KinFace \cite{Siyu2011} & 400           & 600                  & 4                    \\ \cline{2-5} 
                        & KFW-I \cite{lu}     & 1,066         & 1,066                & 4                    \\ \cline{2-5} 
                        & KFW-II \cite{lu}    & 2,000         & 2,000                & 4                    \\ \cline{2-5} 
                        & WVUKin \cite{kohli16}    & 226           & 904                  & 7                    \\ \cline{2-5} 
                        & Family101 \cite{fang2013kinship}    & 607         & 14,816                & 4                    \\ \cline{2-5} 
                        & TSKinFace \cite{qin}    & 2,589         & 2,000                & 4                    \\ \cline{2-5} 
                        & FIW  \cite{robinson2016families}      & 10,700        & 31,000                & 11                    \\ \hline \hline
\multirow{2}{*}{Video}  & UvA-Nemo \cite{uvanemo}   & 152 & 514 & 7 \\ \cline{2-5} 
                        & \textbf{Proposed KIVI}       & \textbf{503}           & \textbf{503}                  & 7                    \\ \hline
\end{tabular}
\end{table}
\endgroup
In 2013, Dibeklioglu et al. \cite{uvanemo} introduced the problem of kinship verification in videos. They computed spatio-temporal facial-dynamics features of high quality videos to model kinship characteristics between two individuals. They used UvA-NEMO Smile database for experimental purposes where smile based dynamic spatio-temporal features and complete local binary patterns on three orthogonal planes  \cite{pfister2011} were extracted and used for kinship verification. This database was initially developed for expression variations and was captured in controlled settings. Recently, Dibeklioglu \cite{dibeklioglu2017visual} introduced visual transformation between the facial appearance of kin for the task of video-based kinship verification.

An important factor in stimulating research in kinship verification is availability of databases.  As shown in Table \ref{tab:kindb}, there are several image-based databases with Family in the Wild (FIW) \cite{robinson2016families} being the largest. Except FIW and WVUkin \cite{kohli16}, other databases focus on 4 key relationships. On the other hand, there is only one existing kinship video database, UvA-Nemo \cite{uvanemo}, which has over 500 videos encompassing 152  individuals and 7 kin-types.    

\subsection{Research Contributions} 
It is interesting to observe that majority of the research on kinship classification has focused on small-scale image databases. However, their applicability also lies in large-scale real-world unconstrained scenarios. It is our assertion that existing algorithms may not be able to scale well to unconstrained scenarios due to challenges such as variations in pose, illumination, and expression. As these covariates are inherently present in videos, they are a natural extension of the image-based kinship verification problem. Videos provide larger sources of information (as shown in Fig. \ref{fig:motivation}) and the spatio-temporal information across different frames can be utilized for learning more complex representations as compared to still images. In this research, we utilize a deep learning framework to learn the spatio-temporal kinship information in videos. We also introduce a new unconstrained video-based kinship face database. The key contributions of this paper are:
\begin{enumerate}

\item A novel Supervised Mixed Norm AutoEncoder (SMNAE) using $\ell_{2,p}$ norm and class-based reconstruction penalty is proposed to learn kinship-specific representation. 

\item A deep learning framework is proposed for kinship verification in unconstrained videos. The proposed multi-stage framework utilizes the spatio-temporal information present in the video frames to verify kinship.
 
\item A new kinship video (KIVI) face database of 503 individuals with \textit{wild} variations due to pose, illumination, occlusion, ethnicity, and expression is introduced. The database consists of 252,804 frames corresponding to seven kin-relations. 

\end{enumerate}



\section{Proposed Kinship Verification in Unconstrained Videos using SMNAE  }
\label{sec:framework}

\begin{figure*}
 \centering
 \includegraphics[width=0.9\linewidth]{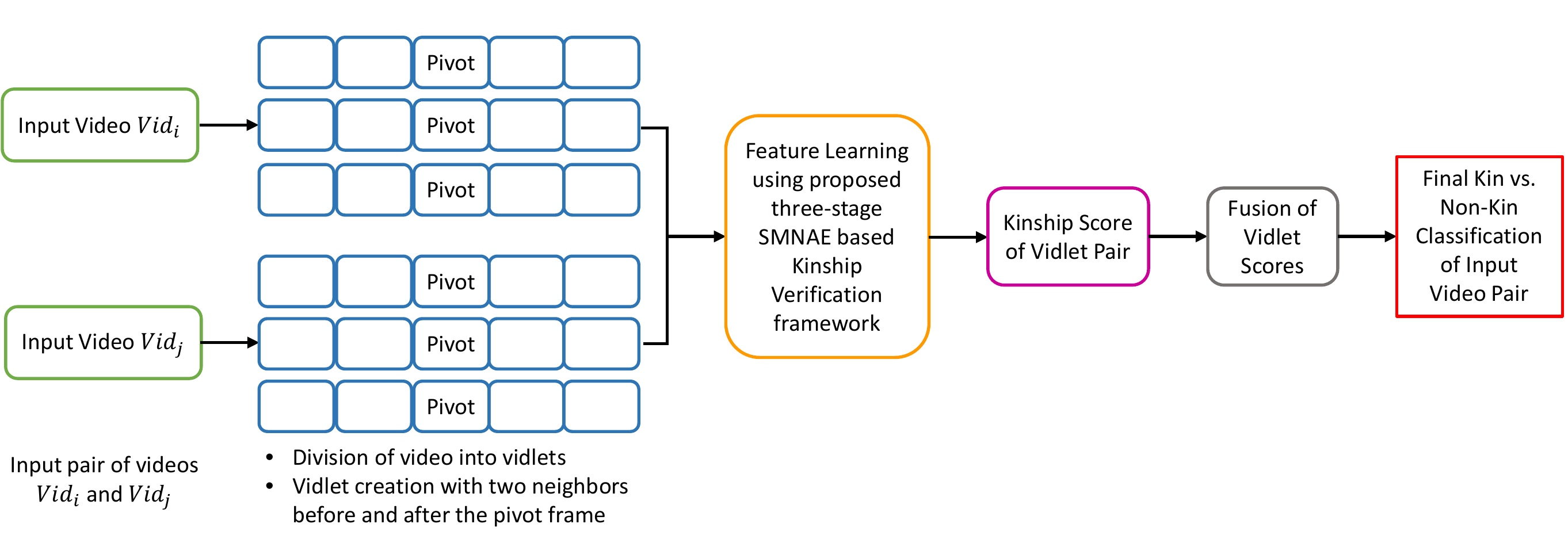}
 \caption{Proposed kinship verification framework: two input videos are divided into non-overlapping vidlets and for every vidlet, SMNAE features are extracted. The final video classification is performed by fusion of all the vidlet pair scores.} 
 \label{fig:broad}
\end{figure*}

In this section, we elaborate the proposed deep learning framework for kinship verification using unconstrained videos. This framework utilizes a novel formulation of autoencoder using $l_{2,p}$  norm and class-based reconstruction error to promote class-based sparsity in the learned weight matrix. The details of the proposed SMNAE and kinship verification framework are described in the following subsections.
\subsection{Supervised Mixed Norm Autoencoder (SMNAE)}

An autoencoder consists of an encoder that maps the input data ($X$) to the latent space and a decoder that maps the learned representation to its reconstruction by minimizing the following loss function: 
\begin{equation}
\argmin_{\mathbf{W},\mathbf{W'}} \parallel \mathbf{X} - \phi ( \mathbf{W}' \phi (\mathbf{W}\mathbf{X}) ) \parallel_{F}^{2} + \lambda R
\label{eq:autoencoder}
\end{equation}

\noindent where, $\mathbf{W}$ is the weight matrix to be learned, $\phi$ is the activation function, $\lambda$ is the regularization constant of the regularizer $R$, and $\parallel \cdot \parallel_{F}^{2}$ denotes the Frobenius norm. Introducing sparsity constraint on the autoencoders forces autoencoders to learn underlying patterns by utilizing only a few hidden nodes. A variety of regularization schemes such as $\ell_{1}$-norm or $\ell_{2}$- norm can be employed on the weights of the autoencoder to promote sparsity. Sparse autoencoders \cite{ng2011sparse} have been proposed which utilize KL divergence to ensure that the average activation of the nodes stays below the pre-defined sparsity parameter. 

Recently, new formulations of supervised autoencoders have been introduced that learn representative features for each class separately. They utilize the class labels to enforce class-wise constraints on the weights of the autoencoders, thus, enforcing discrimination. For learning similarity between faces, Gao et al. \cite{gao2015single} introduced supervised autoencoders using label information. Similarly, Majumdar et al. \cite{majumdar_pami_2016} proposed class-wise sparsity by applying $\ell_{2,1}$ norm on the weight matrix of the autoencoders. However, it has been established that utilizing $\ell_{p}$ norm with $0<p<1$ can find sparser solutions as compared to using $l_1$ norm \cite{chartrand2007exact,chartrand2008iteratively}. Therefore, $\ell_{2,p}$ norm with $0<p<1$ may enforce better sparsity on the weights than $\ell_{2,1}$ norm. Yan et al. \cite{yanl2p} demonstrated the effectiveness of $\ell_{2,p}$ norm for learning local structures of the data distribution for clustering. In this research, we utilize $\ell_{2,p}$ mixed norm for introducing class-wise sparsity on the weights to extort class-specific representative features. 

The loss function $J_{L2P}$ for this autoencoder (referred to as $L_{2,p}$ autoencoder) is defined as:

\begin{align}
\begin{split}
J_{L2P} &= \argmin_{\mathbf{W},\mathbf{W'}} \parallel \mathbf{X} - \phi (\mathbf{W}' \phi (\mathbf{W}\mathbf{X})) \parallel_{F}^{2} \\
& \hspace{2em} + \lambda \sum_{c=1}^{C} \parallel \mathbf{W}\mathbf{X}_{\mathbf{c}} \parallel_{2,p}\\
\end{split}
\label{eq:lpq}
\end{align}

\noindent where, $0 < p < 1$, $X_{c}$ denotes the input data belonging to class $c$, $\text{ and } C$ is the number of classes of the input data. For a weight vector $\mathbf{W}$, $\ell_{2,p}$ norm is computed by taking $\ell_{p}$ norm across the samples in rows of $\mathbf{W}$, $w^{i}$. 
\begin{equation}
\parallel \mathbf{W} \parallel_{2,p} = \left( \sum_{i=1}^{d} \parallel w^{i} \parallel^{p} \right)^{1/p}
\end{equation}
 
\noindent Additionally, we introduce a class-based reconstruction penalty on the autoencoder to encode discriminatory information between classes. The motivation of using this pairwise constraint is to add more discriminating power to the autoencoder. Therefore, we combine $\ell_{2,p}$ norm and pairwise class-based sparsity penalty to propose a new Supervised Mixed Norm Autoencoder. 

Given an input $\mathbf{X}$, containing $N$ pairs of samples for a binary classification problem, the loss function for the proposed SMNAE, $J_{SMNAE}$ is described below:
\begin{equation}
  \begin{split}
 J_{SMNAE} &= \argmin_{\mathbf{W},\mathbf{W'}} \parallel \mathbf{X} - \phi( \mathbf{W}' \mathbf{H}) \parallel_{F}^{2} + \\
 & \lambda \sum_{c=1}^{C} \parallel \mathbf{W}\mathbf{X}_{\mathbf{c}} \parallel_{2,p} + \ \beta ( \Tr( \mathbf{H}^{T}\mathbf{H} \mathbf{L} ) )
   \end{split}
   \label{eq:JSMNAE}
\end{equation}

\noindent where, $\lambda$ and $\beta$ are the regularization constants of respective cost functions, $\Tr$ denotes the trace of the matrix, Laplacian matrix $\mathbf{L}$ is constructed as $\mathbf{L} = \mathbf{D} - \mathbf{M}$, $\mathbf{D}$ is the diagonal matrix formed as $\mathbf{D} = \text{diag}( d_{1}, d_{2}, \ldots, d_{N})$, $d_{i} = \sum_{j=1}^{C}M_{i,j}$, $\mathbf{H} = \phi(\mathbf{W}\mathbf{X})$, and $\mathbf{M}$ is a matrix such that \[  M_{i,j} = \begin{cases}
+1, \text{ \ \ if } x_{i} \text{ and } x_{j} \text{ belong to same class } \\
-1, \text{ \ \ if }  x_{i} \text{ and } x_{j} \text{ belong to different classes }\\
\ \ 0, \text{\ \ \ otherwise }\\
\end{cases} \] 

\begin{figure*}[t]
 \centering
 \includegraphics[width=.8\linewidth]{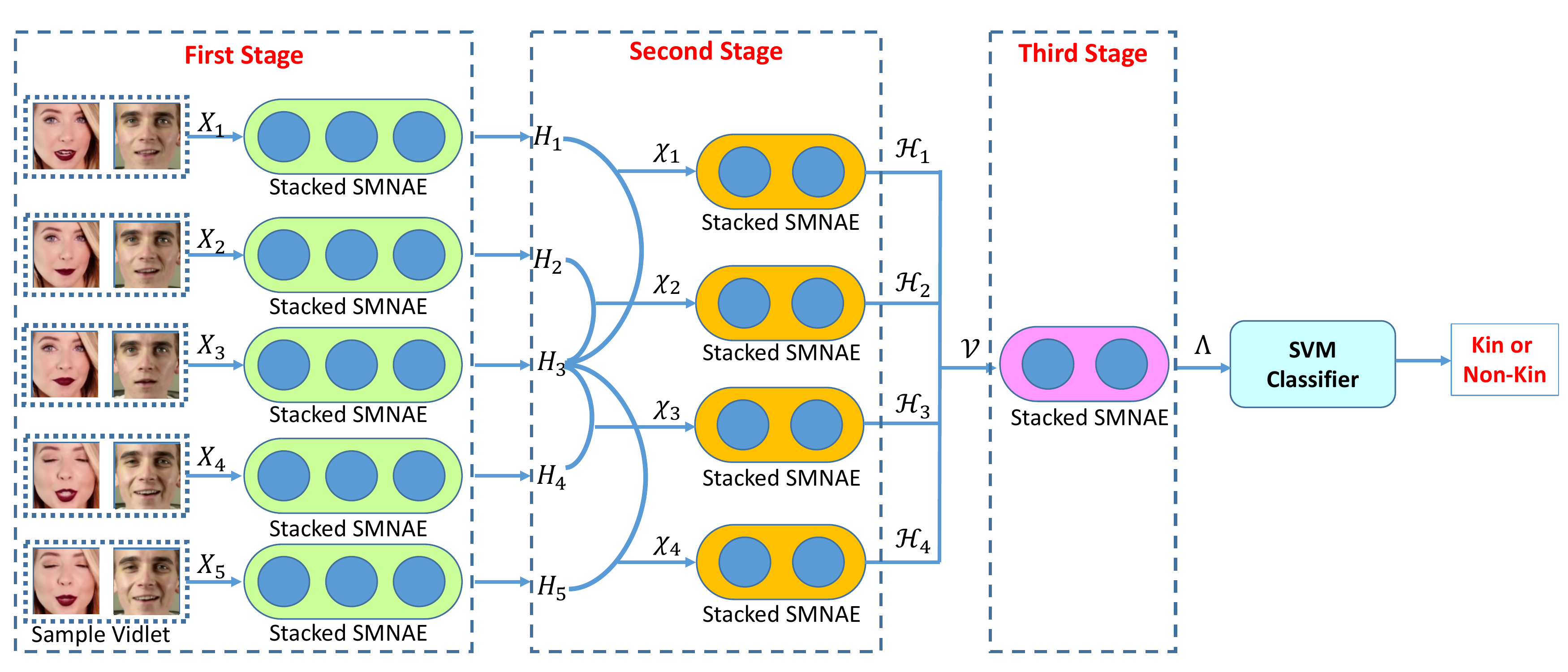}
 \caption{Proposed three-stage kinship verification in unconstrained videos framework by utilizing SMNAE. In the first stage, a pair of videos is split into vidlets which are provided as input to stacked SMNAE. The spatial representations learned from the first stage are concatenated in a pairwise fashion and provided to the second stage stacked SMNAE. The third stage learns the kinship-specific representation of the vidlets using stacked SMNAE to encode the spatio-temporal information which is employed as the input to SVM for kin vs non-kin video classification.} 
 \label{fig:block}
\end{figure*}

The loss function in Eq. \ref{eq:JSMNAE} can be rewritten as:
\begin{equation}
J_{SMNAE} = \ \ J_{1} + \ \lambda  J_{2} + \ \beta  J_{3}  
\end{equation}
where, $J_{1}$ is the first part denoting the traditional autoencoder loss function, $J_{2}$ is the middle term belonging to the $\ell_{2,p}$ norm, and $J_{3}$ is the last term denoting the pairwise class-based discrimination term. Here, $J_{1}$ and $J_{3}$ are convex and smooth, whereas $J_{2}$ is non-convex. This equation can be converted to:
\begin{equation}
\argmin_{\mathbf{W}} J_{SMNAE}(\mathbf{W}) = f(\mathbf{W};\mathbf{X}) +  \lambda \sum_{c=1}^{C} \parallel \mathbf{W}\mathbf{X}_{\mathbf{c}} \parallel_{2,p}
\label{eq:minimization}
\end{equation}

\noindent where, $f(\mathbf{W};\mathbf{X})$ is a convex function consisting of $J_{1}$ and $J_{3}$ and whose gradient is Lipschitz continuous. The proximal gradient algorithm \cite{jenatton2010proximal,nesterov2007gradient} is applied to solve the optimization problem shown in Eq. \ref{eq:minimization}. Beginning with an initial value of $\mathbf{W}_{0}$, $\mathbf{W}$ is updated iteratively using $\mathbf{W}_{1}, \mathbf{W}_{2}, \ldots, \mathbf{W}_{t}$. Thus, the proximal operator equation becomes the following and can then be solved using the Newton's method \cite{Zhang2014}:
\begin{equation}
\mathbf{W}_{t+1} = \argmin_{\mathbf{W}} \frac{1}{2\eta} \parallel \mathbf{W} - \mathbf{A} \parallel_{F}^{2}  + \lambda \sum_{c=1}^{C} \parallel \mathbf{W}\mathbf{X}_{\mathbf{c}} \parallel_{2,p}
\label{eq:proximal}
\end{equation}

\noindent where, $\mathbf{A} = \mathbf{W}_{t} - \eta \nabla f(\mathbf{W_{t}})$ represents the update for $\mathbf{W}_{t}$. Considering linearity, $\nabla f(\mathbf{W_{t}})$ can be represented as
\begin{align}
\begin{split}
\nabla f(\mathbf{W_{t}}) &= \frac{\partial J_{1}}{\partial \mathbf{W_{t}}} + \frac{\partial J_{3}}{\partial \mathbf{W_{t}}} \\
&= -2 \mathbf{X} [ \mathbf{X} - \mathbf{W'}\mathbf{W} \mathbf{X} ] + 2 \mathbf{X} \beta \mathbf{W} \mathbf{X}\mathbf{L}
\end{split}
\end{align}

In this paper, we utilize this proposed formulation of SMNAE to develop a framework for unconstrained video-based kinship verification. The $\ell_{2,p}$ norm on the weights of the autoencoders introduces sparsity for kin and non-kin class individually where the value of $p$ can be tuned to obtain the optimum amount of sparsity in the features. The pairwise constraint terms enforce better discrimination between the kin and non-kin class representations. 

\subsection{SMNAE for Kinship Verification in Unconstrained Videos}
Using SMNAE as the integral representation learning unit, we propose a framework for verifying kinship relation in a pair of videos as shown in Fig. \ref{fig:broad}. Each video in the input pair of videos is decomposed into \textit{vidlets}. A vidlet is defined as a non-overlapping contiguous group of frames with $z$ number of neighboring frames before and after the pivot frame. The center frame of the vidlet is the pivot frame and the  variations across the small temporal neighborhood of the pivot frame are learned. In addition, the kinship-specific spatial information in every frame is encoded. A pair of vidlets is passed as input to the proposed SMNAE based framework. The framework classifies the vidlet pair as kin or non-kin. This process is repeated for every vidlet pair. Fusion of these vidlet scores is performed using sum rule fusion to produce the final classification of the input pair of videos.

Fig. \ref{fig:block} illustrates the detailed architecture of vidlet feature learning and classification in the proposed kinship verification framework. The training of the proposed three-stage SMNAE framework is performed in a modular fashion where the first stage learns the spatial kinship representation of images in each frame. The second stage training learns the pairwise spatio-temporal kinship representation of frames in a vidlet, while the third stage training learns the cumulative vidlet kinship representation, aggregating both the spatial and temporal features. Let $Vid_{i}$ and $Vid_{j}$ be the videos of subjects $i$ and $j$ respectively. The individual frames of the videos of subject $i$ are denoted as $Vid_{i} = [Fr_{i}^{1}, \ldots, Fr_{i}^{n} ]$. The input to the proposed framework is a pair of vidlets from both the videos. The three stages of the proposed framework are discussed in detail below.

\subsubsection{First Stage - Learning Spatial Kinship Representations}

In the first stage of the proposed framework, compact kinship-specific representations are learned from frame-level spatial information. The input pair of videos is treated as a stream of vidlets. Each vidlet comprises frames $[Fr_{i}^{(piv-z)}, \ldots, Fr_{i}^{piv}, \ldots, Fr_{i}^{(piv+z)}]$ and $[Fr_{j}^{(piv-z)}, \ldots, Fr_{j}^{piv}, \ldots, Fr_{j}^{(piv+z)}]$, where $ Fr_{i}^{piv}$ is the pivot frame. The corresponding frames of the pair of vidlets are concatenated and presented as input to stacked SMNAE to learn the spatial representation of kinship in the images. In this first stage,

\begin{itemize}

\item Input: $\mathbb{X} = [Fr_{i}^{k}, Fr_{j}^{k}] \ \forall \ k$ in  $\{1,\ldots,(2*z+1)\}$, and the regularization parameters $\lambda$ and $\beta$.

\item Output: Learned weight matrix $\mathbf{W}$ and $\mathbf{H} = \phi(\mathbf{W}\mathbb{X} )$

\item Loss function ($J_{F}$): 
\begin{align}
\begin{split}
J_{F} &= \argmin_{\mathbf{W},\mathbf{W'}} \ \ \parallel \mathbb{X} - \ \ \phi ( \mathbf{W}' \mathbf{H} )\parallel_{F}^{2} + \\
 & \lambda \sum_{c=1}^{C} \parallel \mathbf{W}\mathbb{X}_{\mathbf{c}} \parallel_{2,p} + \ \beta ( \Tr( \mathbf{H}^{T}\mathbf{H} \mathbf{L} ) ) 
 \label{eq:firststage}
\end{split} 
\end{align}

\end{itemize}

For all three stages, $\mathbf{D} = \text{diag}( d_{1}, d_{2}, \ldots, d_{N})$, $d_{i} = \sum_{j=1}^{N}M_{i,j}$, $\mathbf{L} = \mathbf{D} - \mathbf{M}$, and  $\mathbf{M}$ is created such that \[  M_{i,j} = \begin{cases}
+1, \text{ \ \ if  Vid}_{i} \text{ and Vid}_{j} \text{ belong to kin } \\
-1, \text{ \ \ if Vid}_{i} \text{ and Vid}_{j} \text{ belong to non-kin }\\
\ \ 0, \text{\ \ \ otherwise }\\
\end{cases} \]

\subsubsection{Second Stage - Learning Pairwise Spatio-temporal Kinship Representations in Temporal Neighborhood}

To account for spatial changes across a small temporal neighborhood, a second stacked SMNAE is trained, greedily layer by layer.  The second stage SMNAE is trained to learn and encode variations between a pivot frame and its neighboring frames. These encodings facilitate learning a robust representation between kin pairs across variations due to occlusion, illumination, pose, and expression changes.

The output of first stage is the learned representation of the pair of $(2*z+1)$ frames of the vidlet and is denoted by $H^{k}$, $\forall \ k \text{ in } \{1,\ldots,(2*z+1)\}$. In this stage, the pivot index ($piv$) is chosen and $2*z$ pairwise combinations ($\mathcal{X}$) are formed between the representations of the pivot and neighboring input frames learned from the first stage. In the second stage of the proposed kinship verification framework:

\begin{itemize}
\item Input: $\mathcal{X}^{o} = [ H^{piv} H^{o}] \text{ such that } o \in \{(piv-z),(piv+z)\} \land o \neq piv $ where $H^{k}=\phi(\mathbf{W}\mathbb{X}^{k}) \ \forall \ k \text{ in } \{1,\ldots,(2*z+1)\}$ and the regularization parameters $\alpha$ and $\gamma$.
\item Output: Learned weight matrix $\mathcal{W}$  and $\mathcal{H} = \phi(\mathcal{W}\mathcal{X})$
\item Loss function ($J_{S}$): 
\begin{align}
\begin{split}
J_{S} &= \argmin_{\mathcal{W},\mathcal{W'}} \parallel \mathcal{X} - \phi( \mathcal{W}' \mathcal{H}) \parallel_{F}^{2} + \\
 & \alpha \sum_{c=1}^{C} \parallel \mathcal{W}\mathcal{X}_{c} \parallel_{2,p} + \ \gamma ( \Tr( \mathcal{H}^{T}\mathcal{H} \mathbf{L} ) )
 \end{split}
 \label{eq:secondstage}
\end{align}
\end{itemize}


\subsubsection{Third Stage - Learning Kinship-Specific Vidlet Representations}
After learning the pairwise spatio-temporal representation of the frames in a vidlet in the second stage, the third stage of the proposed SMNAE based framework encodes the cumulative vidlet kinship representation. It aggregates the spatial and temporal representations to form the final compact representation for the vidlet. This learned representation for a subject is hierarchical in nature, including representation of a single frame, pair of frames, and group of frames. 

For the final stage of the proposed kinship verification:
\begin{itemize}
\item Input: $\mathcal{V} = [\mathcal{H}^{1}, \ldots, \mathcal{H}^{(2*z)}]$ and the regularization parameters $\zeta$ and $\kappa$. 
\item Output: Learned weight matrix $\mathcal{G}$ and $\Lambda = \phi(\mathcal{G}\mathcal{V})$
\item Loss function ($J_{T}$): 
\begin{align}
\begin{split}
J_{T} &= \argmin_{\mathcal{G},\mathcal{G'}} \parallel \mathcal{V} - \phi (\mathcal{G}' \Lambda ) \parallel_{F}^{2} + \\
 & \zeta \sum_{c=1}^{C} \parallel \mathcal{G}\mathcal{V}_{c} \parallel_{2,p} + \ \kappa ( \Tr( \Lambda^{T}\Lambda \mathbf{L} ) )
\end{split}
\label{eq:thirdstage}
\end{align}
\end{itemize}

For a pair of test videos, similar steps are followed where the given pair of videos is first converted into vidlets. If two videos have different lengths, the frames of the shorter video are iteratively cycled to match the number of frames of the longer video. The vidlets are passed through the three stages of the proposed video-based kinship verification framework. The representation of all the vidlets is extracted and provided to the SVM for classification.

\subsubsection{Classification}
The final spatio-temporal representation of streams of vidlets is obtained using the third stage learned weight matrix, $\mathcal{G}$. This learned representation $\Lambda$ for each vidlet is provided to a Support Vector Machine (SVM) \cite{cortes1995support} with radial basis function kernel for computing the decision boundary between the kin and non-kin class. The final classification score for the input pair of videos is a sum of the probability scores of each vidlet obtained as output from the SVM. 

\subsection{Implementation Details}
Every video is pre-processed and contiguous frames are extracted. Face detection and alignment is performed on each frame of the video to obtain an image of size $128 \times 128$. This is followed by vidlet creation. For experimental purposes, three different values of $z$ (pivot's neighboring frames) are analyzed with $z=1, 2, \text{and } 3$. 

 In the first stage of the proposed kinship verification framework, three Supervised Mixed Norm autoencoders (SMNAEs) are stacked together, each containing 8192, 4096, and 2048 hidden nodes in their layers, respectively. The spatial representation learned from the first stage of the proposed framework is used to extract pairwise frame representation. This is used as input to the second stage of the proposed framework.

In the second stage, the third frame pair in the vidlet represents the pivot with $z=2$. The spatial kinship representations learned from the first stage are concatenated in a pairwise fashion and sent to the second stage stacked SMNAEs. The second stage consists of two stacked SMNAE autoencoders, each containing 2304 and 1024 hidden nodes. Finally, all the pairwise spatio-temporal representations learned from the second stage of the proposed kinship verification framework are concatenated to form the input to the stacked autoencoder in the third stage. 

The final stage learns the kinship-specific representation of the vidlets which aggregates the spatial and temporal information. The stacked SMNAE autoencoder used in this stage consists of two autoencoders, each containing 3072 and 2048 hidden nodes. A vector of size 2048 is extracted for each vidlet and is used as input to the SVM classifier. The score for the complete video is calculated as the sum of the probability scores of all the vidlets from the binary SVM classifier.

\begin{figure}[!t]
 \centering
 \includegraphics[width=0.99\linewidth]{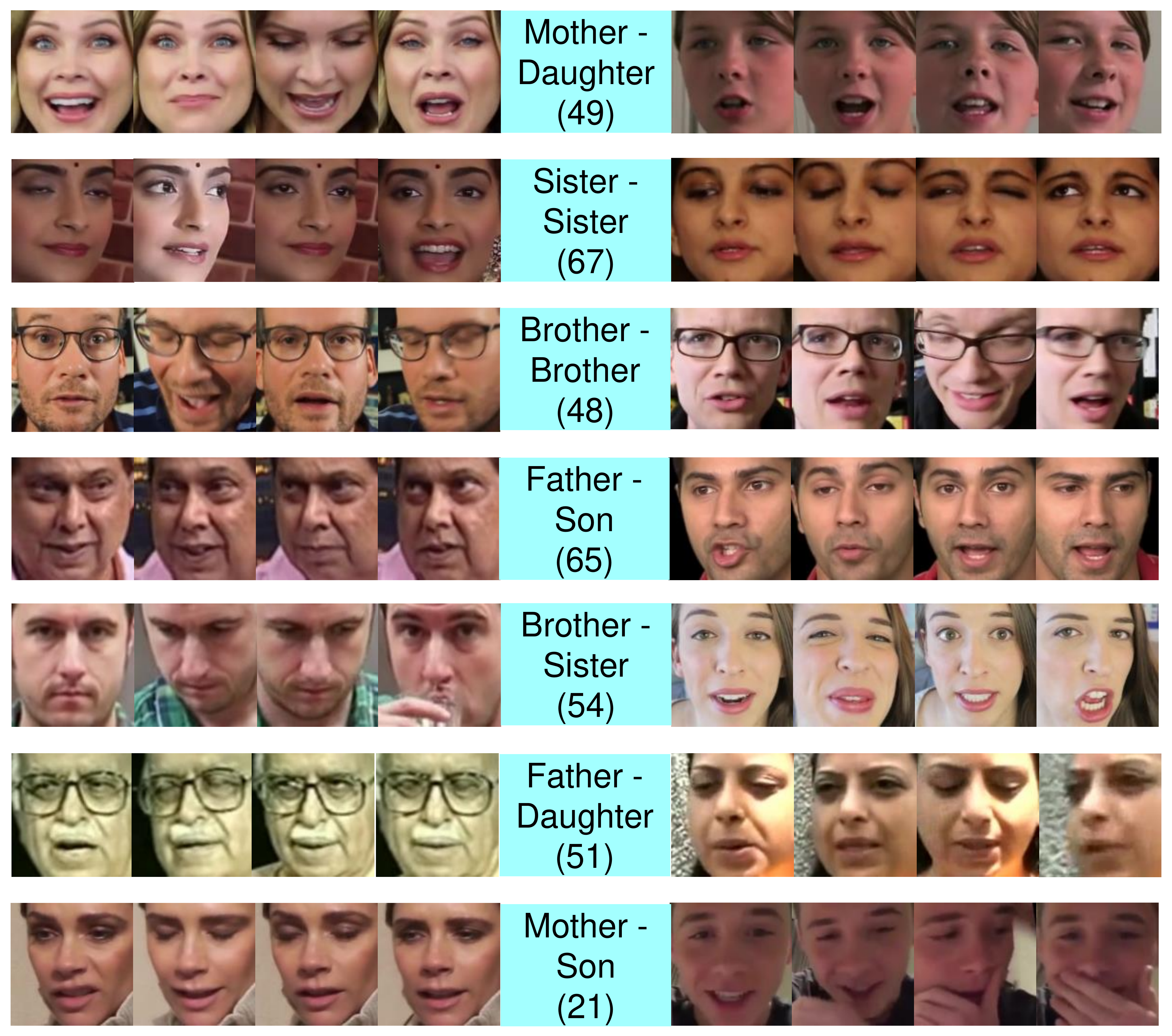}
 \caption{Sample kin-pair subjects from Kinship Video (KIVI) face database comprising of seven kin-relations. The number of respective kin pairs present in each kin-relations category in KIVI database is also indicated.}
 \label{fig:kiviSamples}
\end{figure}

\section{Kinship Video (KIVI) Database}
\label{sec:db}
Existing kinship video database, UvA-NEMO Smile \cite{uvanemo} database was initially created for capturing smile and expression variations. However, it was captured in controlled settings with limited real-world variations and is not suitable for kinship verification in uncontrolled videos. Therefore, to include realistic variations observed in the \textit{wild} we collected the largest unconstrained kinship video (KIVI) face database. It consists of video sequences of 503 individuals forming a total of 355 positive kin pairs from 211 families. For creating the proposed KIVI database, popular celebrity families are selected and the ground truth is annotated manually by validating the kinship information from the Internet. For each subject in the database, there is one corresponding video. The average video duration is 18.78 seconds with the average frame rate of 26.79 frames per second (fps) and the total number of still frames in the database is over 250,000. As illustrated in Fig. \ref{fig:kiviSamples} the database includes seven kin-relations..

\begingroup
\renewcommand*{\arraystretch}{1.1}
\begin{table}[t]
\centering
\caption{Characteristics of the proposed KIVI database.}
\label{tab:kivi}
\begin{tabular}{|l|L{5.05cm}|}
\hline
No. of Individuals              & 503 \\ \hline
No. of Kin Relations            & 7 (Brother-Brother, Sister-Sister, Brother-Sister, Father-Son, Father-Daughter, Mother-Son, and Mother-Daughter)   \\ \hline
No. of Families					& 211 \\ \hline
Ethnicity of Families			& African-American (9), Asian (74), and Caucasian (128)\\ \hline
Average Video Duration          & 18.78 seconds   \\ \hline
No. of Still Frames             & 252,804    \\ \hline
Average Frame Rate              & 26.79 fps   \\ \hline
No. of Video Kin Pairs & 355 Positive and 355 Negative  \\ \hline
No. of Image Kin Pairs & 212,018 Positive and 212,018 Negative \\ \hline
\end{tabular}
\vspace{-3mm}
\end{table}
\endgroup


The database contains videos with variations in illumination, pose, occlusion, ethnicity, and expression collected from the Internet. It has multi-ethnic kin pairs of Asian (74), Caucasian (128), and African-American (9) ethnicity. The characteristics of the KIVI database are summarized in Table \ref{tab:kivi}. The KIVI database will be publicly available to the research community at \url{http://iab-rubric.org/resources/KIVI.html}. For performance evaluation, fixed unseen training and testing partitions of KIVI database are created with non-overlapping subjects. 214 kin subject pairs are chosen at random as the test partition while the remaining kin pairs form the train partition. An equal number of non-kin pairs are added in both the partitions.






\section{Experimental Evaluation} 
\label{sec:eval}

Experimental evaluation of the proposed kinship verification framework in unconstrained videos is performed using the training and testing partitions of the KIVI database. For comparison, Deep+Shallow algorithm proposed by Boutellaa et al. \cite{boutellaa_2016} for kinship verification in videos is utilized. They proposed a fusion of VGG-Face and textural features (LBPTOP \cite{zhao2007dynamic}, LPQTOP, and BSIFTOP) and applied SVM for final classification. Additionally, comparative analysis is performed using existing image-based kinship verification algorithms. It is to be noted that these algorithms are developed to operate on still face images and not videos. Therefore, these existing algorithms are applied to each frame followed by score-level fusion of the frame-wise scores to produce the final classification of the video pair. The equal error rate (EER) is calculated for each experiment and the kinship verification accuracy is reported for all experiments as $(100-\text{EER})$\%. The following image-based kinship verification algorithms are used for comparison purposes and the source codes have been obtained from the authors directly except \cite{wuusefulness}. 

\begin{itemize}

\item Neighborhood repulsed metric learning (NRML) \cite{lu},
\item Chromaticity-based approach \cite{comments2015},
\item Color space approach \cite{wuusefulness},
\item VGG-Face \cite{parkhi2015deep},
\item Stacked Denoising Autoencoder (SDAE) with Neural Network (NN) and Support Vector Machine (SVM),
\item Kinship verification via representation learning - filtered contractive deep belief networks (KVRL-\textit{fc}DBN) \cite{kohli16}.
\end{itemize}

\subsection{Video-based Kinship Verification Results}

\subsubsection{Experimental Protocol}
The objective of the video-based protocol is to compute the kinship verification performance of the proposed framework on a pair of input videos of the KIVI database. The experiments are performed with 40\%-60\% randomly created train-test partitions. There is no overlap between subjects or families in the training and testing folds. The test set consists of 214 positive kin pair videos while the train set contains 141 kin pair videos with an equal number of non-kin pair videos in each set. The entire video is processed and classified as kin or non-kin. The performance of existing image-based techniques and the proposed video-based kinship framework using Supervised Mixed Norm autoencoders (SMNAEs) are evaluated on the test set of the KIVI database. As stated earlier, existing image-based kinship verification algorithms are applied to each frame which is followed by score-level fusion of the frame scores to generate the final kin classification for the video.

\subsubsection{Experimental Results}
\begin{figure}[t]
 \centering
  \begin{subfigure}{0.9\linewidth}
    \centering
    \includegraphics[width=1\linewidth]{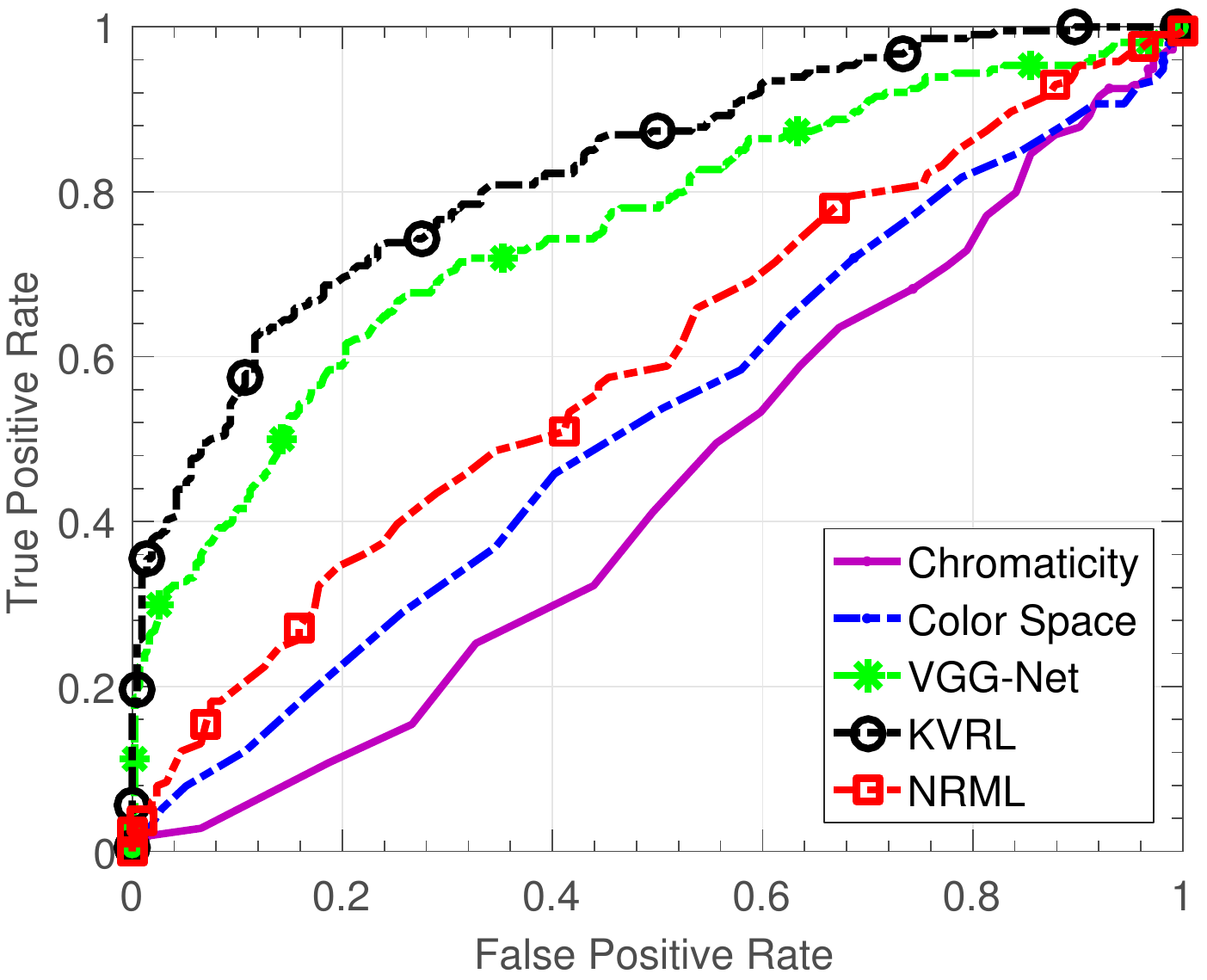}
    \caption{}
    \label{fig:roc_video}
  \end{subfigure} 
\begin{subfigure}{0.9\linewidth}
    \centering
    \includegraphics[width=1\linewidth]{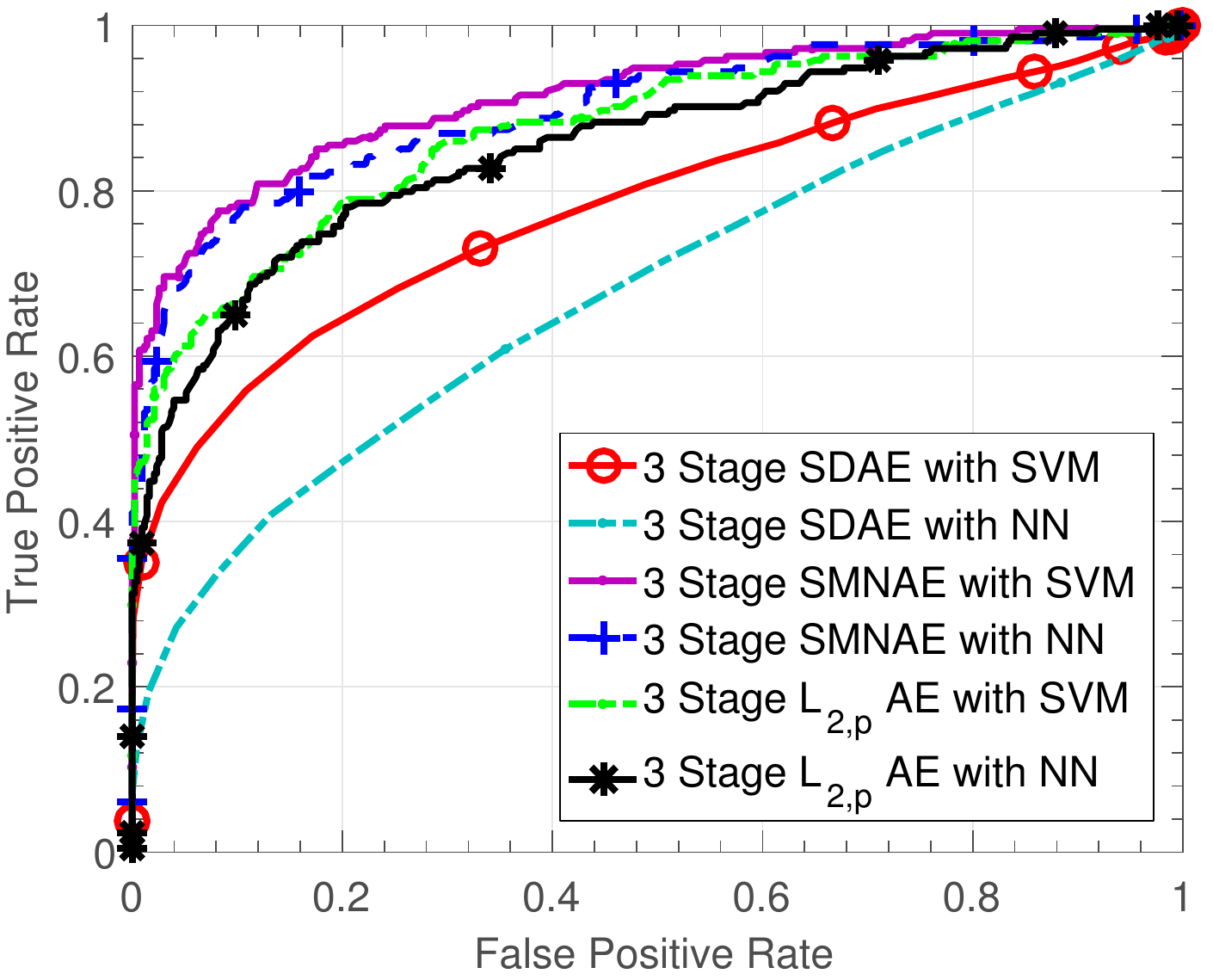}
    \caption{}  
    \label{fig:roc_kivv}
  \end{subfigure}       
  \caption{Video-based kinship verification performance of (a) existing image-based algorithms using frame level aggregation and (b) proposed framework using 3-Stage existing SDAE, 3-Stage $\ell_{2,p}$ AE with mixed norm regularization, and the proposed 3-Stage SMNAEs on the KIVI database.}
  \label{fig:combined_video_roc}
\end{figure}

\begin{figure}[t]
 \centering
    \includegraphics[width=0.75\linewidth]{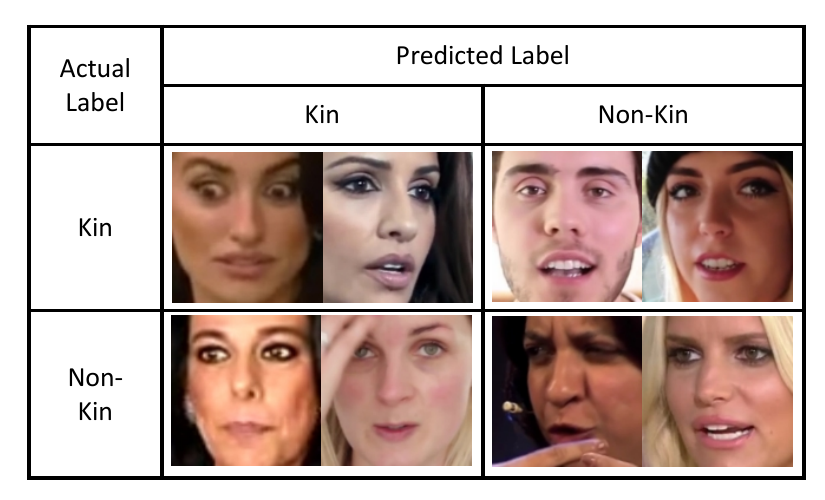}
    \caption{Sample results of the proposed SMNAE framework in the form of a confusion matrix.}
    \label{fig:confMat}
\end{figure}

The results for video-based kinship verification using the proposed framework and existing image-based kinship verification algorithms are  shown in Table \ref{tab:videowise} and Fig. \ref{fig:combined_video_roc}. The proposed framework with SMNAE and SVM yields the highest kinship verification accuracy of 83.18\%. Sample video pairs which are correctly and incorrectly predicted by the proposed framework are shown in Fig \ref{fig:confMat}. The detailed analysis of the results is described below.

\begingroup
\renewcommand*{\arraystretch}{1.2}
\begin{table}[t]
	\centering
	\caption{Video-based kinship verification accuracy (\%) of existing image-based and proposed kinship verification algorithms on KIVI face database.}
	\label{tab:videowise}
	\begin{tabular}{|c|L{5cm}|C{1.7cm}|}
		\hline
\multirow{8}{*}[-1.1em]{\begin{turn}{90}Existing\end{turn}} 	&	\textbf{Kinship Verification Algorithm} & \textbf{Accuracy (\%)} \\ \hline

	&	Chromaticity based approach \cite{comments2015} 	& 43.98  \\\cline{2-3}
	&	Color space approach \cite{wuusefulness} & 52.58  \\\cline{2-3}
    &  Deep+Shallow \cite{boutellaa_2016} & 54.46 \\ \cline{2-3}	
    &	NRML \cite{lu}				& 55.61 \\ \cline{2-3}
	&	SDAE Autoencoder with NN  & 64.48 \\\cline{2-3}
	&	SDAE Autoencoder with SVM & 66.88 \\\cline{2-3}
	&	VGG-Face \cite{parkhi2015deep}  & 70.09\\\cline{2-3}
	&	KVRL-\textit{fc}DBN \cite{kohli16}  & 79.91 \\\hline
\multirow{4}{*}{\begin{turn}{90}Proposed\end{turn}} 	&	$L_{2,p}$ Autoencoder with NN &	78.51	\\\cline{2-3}
	&	$L_{2,p}$ Autoencoder with SVM  & 79.21	\\\cline{2-3}
	&	SMNAE based framework with NN  & 81.78 \\\cline{2-3}
	&	\textbf{SMNAE based framework with SVM} &	\textbf{83.18}	\\\hline
	\end{tabular}
\end{table}
\endgroup

\vspace{6pt}
\noindent \textbf{Comparison with existing approaches:} From Table \ref{tab:videowise} and Fig. \ref{fig:combined_video_roc}, it is observed that the proposed multi-stage kinship verification in videos framework consisting of SMNAE with SVM as the classifier outperforms existing approaches with 83.18\% video-based kinship verification accuracy. It surpasses the accuracy of existing video-based kinship verification algorithm \cite{boutellaa_2016} by over 30\%. 
It is to be noted that the proposed kinship verification framework with SMNAE achieves 3.27\% higher accuracy as compared to the state-of-the-art algorithm for image-based KVRL-\textit{fc}DBN kinship verification algorithm \cite{kohli16}. This indicates that simple aggregation of frame-wise scores may not encode the spatio-temporal information present in videos. Hence, this highlights the need for developing video-specific kinship verification algorithms. 

\noindent \textbf{Effect of autoencoders in the proposed kinship verification framework:} To examine the efficacy of different autoencoders as the representation learning unit in the video based kinship verification framework, the same experiments are performed using traditional stacked denoising autoencoders (SDAE) and $L_{2,p}$ autoencoders in the proposed framework. As observed in Table \ref{tab:videowise}, $L_{2,p}$ autoencoder with SVM yields 12.33\% higher classification accuracy as compared to traditional stacked denoising autoencoders. However, it is to be noted that SMNAE demonstrates around 4\% higher performance than $L_{2,p}$ autoencoder for video-based kinship verification. The increase in performance of SMNAE as compared to $L_{2,p}$ autoencoder illustrates the efficacy of the pairwise class-based reconstructive penalty term in the loss function of SMNAE for verifying kin pair videos.

\begin{figure}
 \centering
 \includegraphics[width=0.75\linewidth]{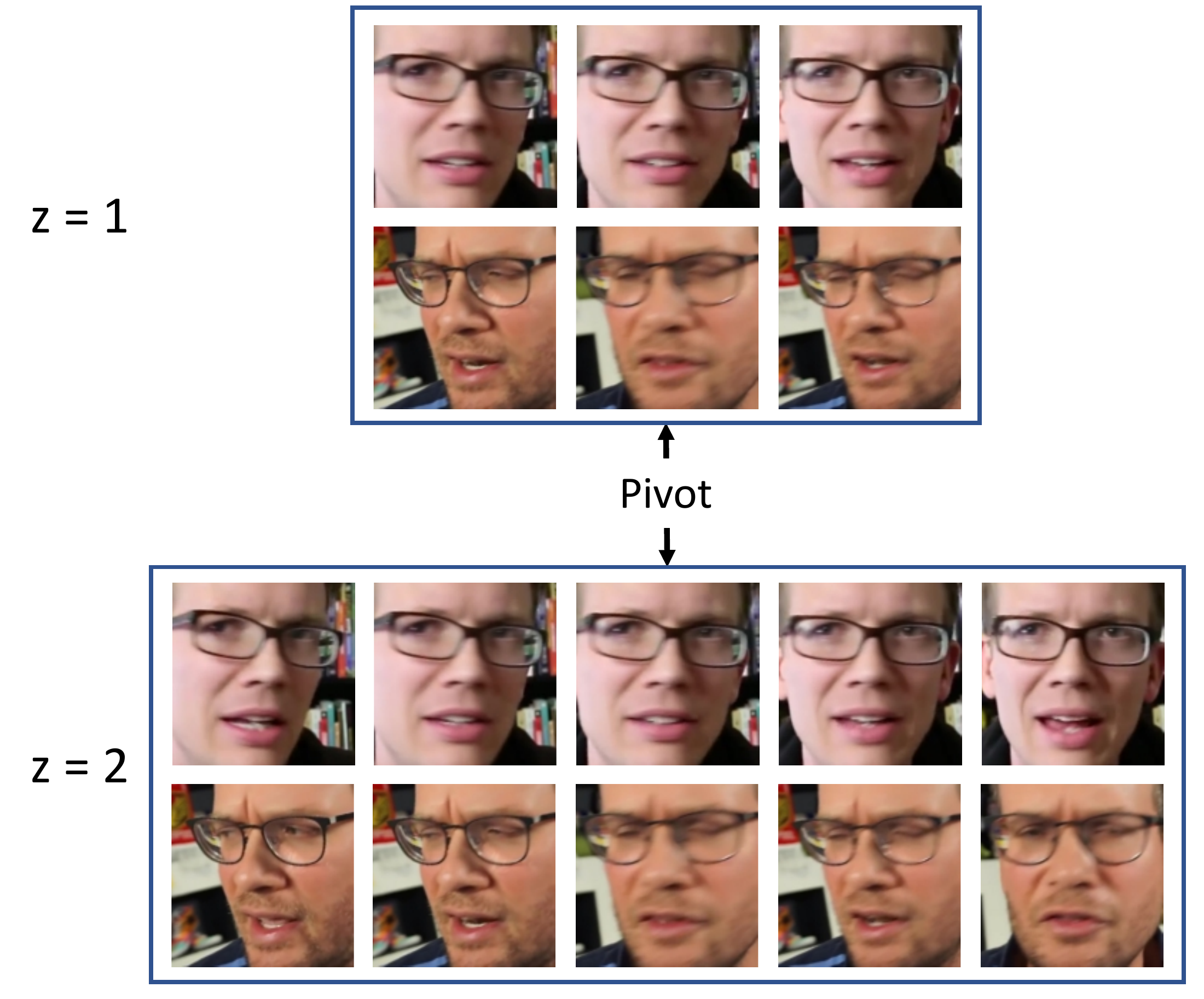}
 \caption{Variations across the pivot frame with respect to number of neighbors ($z$). }
 \label{fig:zexample}
\end{figure}



\vspace{6pt}

\noindent \textbf{Effect of number of neighboring frames ($z$):} The number of neighboring frames ($z$) to be considered with respect to a pivot frame in a vidlet is a crucial parameter. As shown in Fig. \ref{fig:zexample}, the number of neighboring frames selected in the temporal neighborhood affects the content in different frames thereby influencing the learned spatio-temporal representation. Therefore, we next analyze the effect of $z$ in the proposed SMNAE based kinship verification framework. 

The results are computed with $z = 1$ to $z = 3$. It is to be noted that $z = 0$ denotes that there are no neighbors for the pivot frame and hence, is same as the traditional image-based kinship verification experiment. The proposed framework yields 83.18\% kinship verification accuracy with $z = 2$ as compared to 82.24\% with $z = 1$ and 80.14\% with $z = 3$. As shown in Fig. \ref{fig:zexample}, it can be inferred that when the temporal neighborhood is very small, consisting of just one neighboring frame on both sides of the pivot frame ($z = 1$), the encoded spatio-temporal yields lower performance as compared to $z=2$. With $z=3$, the total number of contiguous frames including the pivot frame is 7. As the number of frame increases, the temporal representation is not accurately captured because the frames are not in close proximity of the pivot frame and this decreases the kinship classification performance. Therefore, this experiment illustrates that the best value for $z$ is 2 for the KIVI database.

\vspace{6pt}

\noindent \textbf{Effect of sparsity norm ($p$):} For determining suitable value of $p$ in $\ell_{2,p}$ norm term in Eqs. \ref{eq:firststage}, \ref{eq:secondstage}, and \ref{eq:thirdstage}, comparative analysis is performed with different values of $p$ (0.2, 0.4, 0.6, 0.8, and 1). It is observed that the best performance of 83.18\% is achieved when $p=0.8$ as compared to 77.10\% ($p=0.2$), 78.50\% ($p=0.4$), 78.97\% ($p=0.6$), and 81.13\% ($p=1$). It is noted that smaller values of $p$ produce lower kinship verification accuracy. This may suggest that with smaller values of $p$, the learned network becomes too sparse and hence, encoding inadequate information for the classifier.  However, it is to be noted that different values of $p$ can produce different performance on other databases.


\begingroup
\renewcommand*{\arraystretch}{1.1}
\begin{table}[!t]
\centering
\caption{Video-based kinship verification performance on the seven kin-relations in the KIVI database using the proposed SMNAE based framework.}
\label{tab:kintype}
\begin{tabular}{|L{3.5cm}|C{3cm}|}
\hline
\textbf{Kin Relationship} & \textbf{Accuracy (\%)} \\ \hline
Mother-Daughter & 92.31 \\ \hline
Sister-Sister & 83.64 \\ \hline
Brother-Sister & 82.86 \\ \hline
Father-Daughter & 81.82 \\ \hline
Brother-Brother & 81.25 \\ \hline
Father-Son & 80.00 \\ \hline
Mother-Son & 77.78 \\ \hline
\end{tabular}
\end{table}
\endgroup

\vspace{6pt}
\noindent \textbf{Role of classifier for final kin vs non-kin decision:} The effect of classifier after the third stage of the proposed kinship verification framework is analyzed. The proposed SMNAE based framework with neural network (NN) classifier yields 1.4\% lower accuracy as compared to the performance of SVM with the proposed framework. This result illustrates the effectiveness of SVM in learning the kin vs non-kin classification boundary as compared to the neural network.

\vspace{6pt}

\noindent \textbf{Effect of kin-relation:} Experimental evaluation is performed on all seven types of kin-relation in the input kin pair video. The classification accuracy of the proposed kinship verification framework for different kin-relations is shown in Table \ref{tab:kintype}. It is observed that mother-son pair is the most difficult to detect while mother-daughter video pairs are verified with the highest accuracy of 92.31\%. It has been ascertained that humans are able to verify kin pairs easily if there is at least one female subject in the pair \cite{kaminski2009human}. One reason postulated for this phenomenon is higher degree of variation in facial images of males. A similar trend is noticed here as brother-brother and father-son kin pair videos have lower kinship verification accuracy as compared to other relationships due to the absence of a female subject. However, mother-son relationship kin pairs contradict this pattern and one reason for that can be the fewer number of samples of this relation as compared to other relationships in the KIVI database. 

\vspace{6pt}

\noindent \textbf{Role of fusion of vidlets:} An input video pair of two subjects produces many vidlets. Therefore, in the proposed kinship verification framework, the final classification score for the input pair of videos is a sum of the probability scores of each vidlet obtained as output from the SVM. For comparative analysis, the performance of the proposed framework is computed by applying the max rule on the probability scores. In the max rule, the vidlet pair with the highest probability score is chosen for providing the final classification. Video-based kinship verification accuracy of 80.37\% is obtained with the max rule as compared to 83.18\% by using the sum rule. In sum rule fusion, all the vidlet pairs contribute equally and are combined to produce the final decision. The performance is higher as compared to max rule fusion where only one pair is selected to produce the decision. This demonstrates that it is beneficial to integrate decisions from all the vidlet pairs to produce the final classification of kin or non-kin.



\subsection{Image-based Kinship Verification Results}
The proposed SMNAE based framework achieves state-of-the-art kinship verification accuracy in videos. However, in certain scenarios images may be the only source of information due to unavailability of videos. In such scenarios, kinship can only be determined using image pair of the subjects. The performance of the proposed kinship verification framework on images is reported below. 

\subsubsection{Experimental Protocol}
This protocol is created to evaluate the performance of the proposed kinship verification framework at frame-level (image-level) without taking into account the temporal information present in videos. Existing image-based kinship verification algorithms are applied to each frame. Corresponding frames of the input pair of video are extracted and kinship verification is performed on a per-frame basis. On average, the test set consists of 125,708 pairs of positive kin frames while the train partition contains 86,310 pairs of positive kin pair frames. An equal number of non-kin frame pairs are added in both test and training partitions.

\subsubsection{Experimental Results}

\begin{figure}[t]
\centering
	\begin{subfigure}[t]{1\linewidth}
    \centering
    \includegraphics[width=0.9\linewidth]{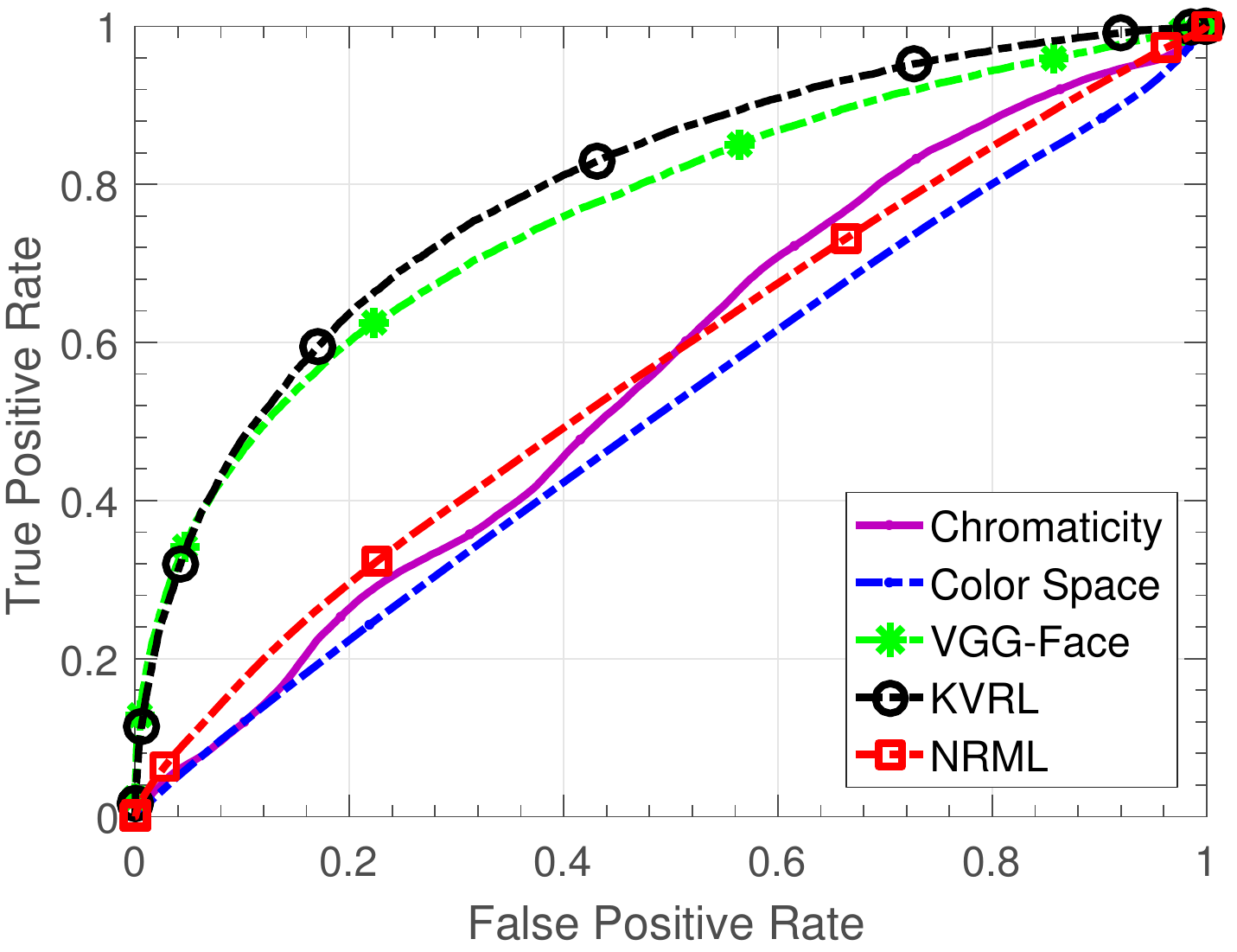}
    \caption{}  
    \label{fig:roc_frame}
  \end{subfigure} \hfill
  ~
  \begin{subfigure}[t]{1\linewidth}
    \centering
    \includegraphics[width=0.9\linewidth]{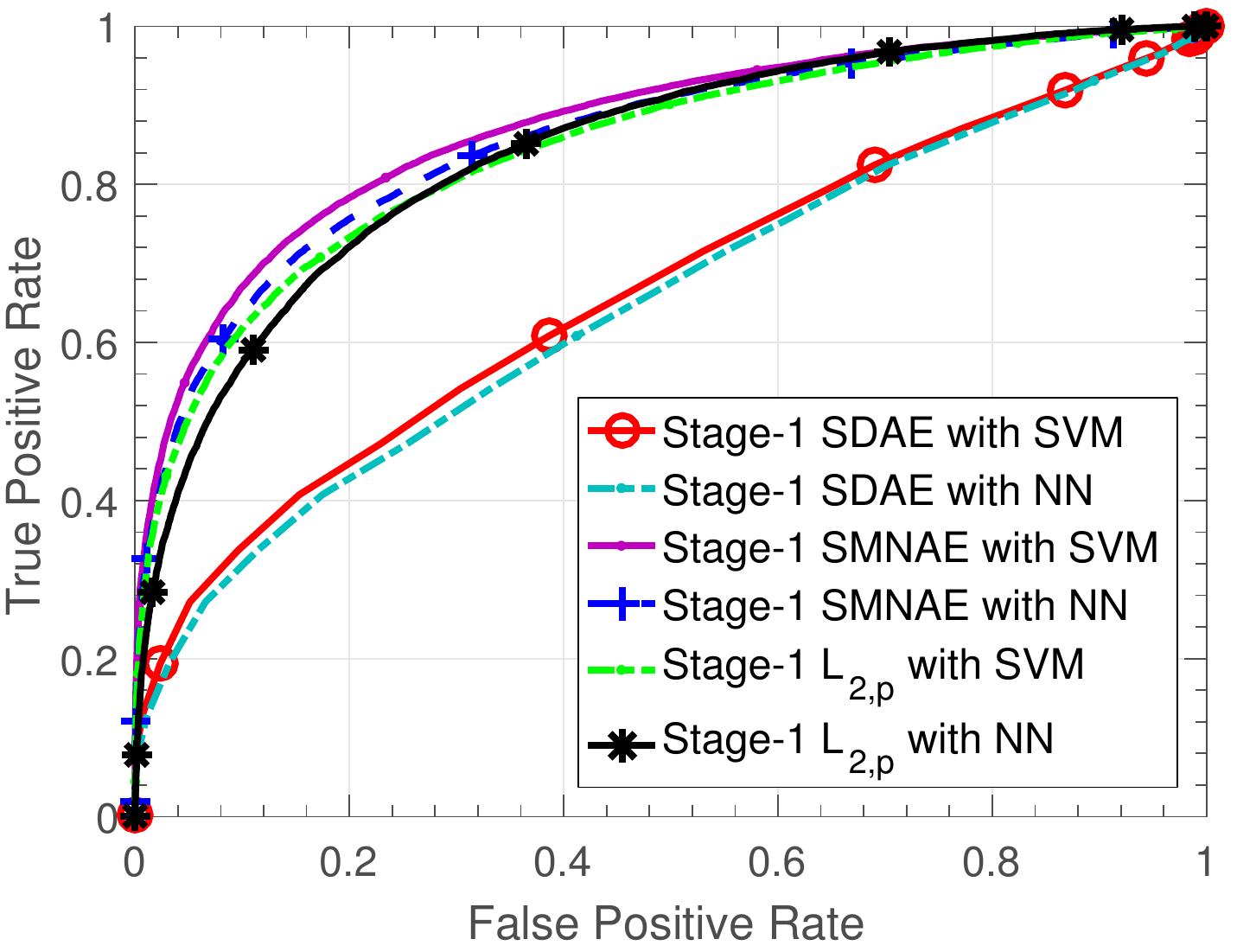}
    \caption{}
    \label{fig:roc_stage1_ae_frame}
  \end{subfigure} 
  \caption{Image-based kinship verification performance of (a) existing image-based algorithms (b) first stage of SDAE, $L_{2,p}$ AE, and proposed SMNAE on still frames of KIVI database.} 
  \label{fig:roc_stage1}
\end{figure}

The kinship verification performance is evaluated on per-frame basis and the results are reported in Table \ref{tab:framewise} and Fig. \ref{fig:roc_stage1}. 

\begingroup
\renewcommand*{\arraystretch}{1.2}
\begin{table}[t]
\centering
\caption{Image-based kinship verification accuracy of existing image-based and proposed kinship verification algorithms on still frames of the KIVI face database.}
\label{tab:framewise}
\begin{tabular}{|l|L{5.2cm}|C{1.75cm}|}
\hline
& \textbf{Kinship Verification Algorithm} & \textbf{Accuracy (\%)} \\ \hline
\multirow{7}{*}{\begin{turn}{90}Existing\end{turn}} 
& Color space approach \cite{wuusefulness} & 51.18   \\ \cline{2-3}
& Chromaticity based approach \cite{comments2015}	& 54.00   \\ \cline{2-3}
& NRML \cite{lu}	& 54.47  \\ \cline{2-3}
& SDAE Autoencoder with NN  & 58.77  \\ \cline{2-3}
& SDAE Autoencoder with SVM  & 61.33  \\ \cline{2-3}
& VGG-Face \cite{parkhi2015deep}  & 69.32 \\ \cline{2-3}
& KVRL-\textit{fc}DBN \cite{kohli16}  & 77.97  \\\hline
\multirow{4}{*}{\begin{turn}{90}Proposed\end{turn}} & First stage $L_{2,p}$ Autoencoder with NN& 76.20	\\\cline{2-3}
& {First stage $L_{2,p}$ Autoencoder with SVM}  & 76.48 \\\cline{2-3}
& {First stage SMNAE with NN}  & 77.41	\\\cline{2-3}
& \textbf{First stage SMNAE with SVM} & \textbf{79.09} \\\hline
\end{tabular}
\end{table}
\endgroup

\vspace{6pt}

\noindent \textbf{Comparison with existing approaches:} The performance of the proposed video-based kinship verification framework is evaluated on images using the first stage of the kinship verification framework as this stage of the framework is intended to encode the spatial features irrespective of the temporal information. It is observed that the first stage of stacked SMNAE with SVM classifier yields the highest classification accuracy of 79.09\% on the KIVI database. Furthermore, it is observed that the state-of-the-art algorithm for image-based kinship verification, KVRL-\textit{fc}DBN \cite{kohli16} yields kinship verification accuracy of 77.97\%. In \cite{kohli16}, the authors have reported that KVRL-\textit{fc}DBN achieves more than 90\% classification accuracy on five publicly available kinship image databases. This decrease in the performance of KVRL-\textit{fc}DBN \cite{kohli16} on KIVI database demonstrates the challenging nature of the database due to real-world variations in pose, illumination, occlusion, ethnicity, and expressions. 
\subsection{Performance on Existing Databases}

To demonstrate the efficacy of the proposed SMNAE framework on existing publicly available kinship databases, five image-based and one video-based kinship databases are selected. The experimental evaluation on these databases is performed using the pre-defined protocols described in the respective papers. Comparative analysis on these databases is performed using existing kinship verification algorithms and the results are summarized in Table \ref{tab:otherImgDB}.

The first stage of the proposed SMNAE based framework, signifying the scenario when $z=0$, is fine-tuned and evaluated on the following five publicly available image-based kinship databases: (i) Cornell \cite{Fang2010}, (ii) UB KinFace \cite{Siyu2011}, (iii) KinFaceW (KFW)-I \cite{lu}, (iv) KFW-II \cite{lu}, and (v) WVU Kin \cite{kohli16}. The performance of the proposed SMNAE is compared with the following existing kinship verification algorithms: (i) NRML \cite{lu}, (ii) PDFL \cite{Yan2014}, (iii) DDMML \cite{lu2017discriminative}, (iv) VTCL \cite{dibeklioglu2017visual}, and (v) KVRL-\textit{fc}DBN \cite{kohli16}. The analysis of the results reported in Table \ref{tab:otherImgDB} illustrates that the proposed SMNAE outperforms other existing algorithms on all five image databases by 0.8\% to 4.9\% on the task of kinship verification.

The performance of the proposed three-stage SMNAE based framework is also evaluated on the UvA-Nemo Smile dataset \cite{uvanemo}. This video dataset consists of 95 subjects having a kin relationship. Following the protocol described by Dibeklioglu et al. \cite{uvanemo}, the proposed SMNAE framework achieves 96.07\% kinship verification accuracy. The three-stage SMNAE framework which encodes spatio-temporal kinship-specific representation outperforms the visual transformation aided technique \cite{dibeklioglu2017visual} by 2.42\%.

  \begingroup
\renewcommand*{\arraystretch}{1}
\begin{table}[!t]
\centering
\caption{Kinship verification accuracy (\%) of kinship classification algorithms on existing databases. Results of existing algorithms are directly reported from the original papers.}
\label{tab:otherImgDB}
\begin{tabular}{|L{1.12cm}|C{0.85cm}|C{0.825cm}|c|c|C{0.675cm}|C{0.8cm}|}
\hline
\multicolumn{1}{|c|}{}                & \multicolumn{5}{c|}{\textbf{Image}}       & \textbf{Video}   \\ \cline{2-7}
\textbf{Method}    & Cornell & UBKin & KFWI & KFWII & WVU Kin & UvA-Nemo \\ \hline
NRML  &  - & - & 69.90 & 76.50 & -  & - \\ \hline
PDFL  & 71.90 & 67.30 & 70.10 & 77.00 & - & - \\ \hline
DDMML & - & - & 83.50 & 84.30 & - & - \\ \hline
VTCL  & - & - & 80.50 & 82.30 & - & 93.65 \\ \hline
KVRL \textit{fc}DBN & 89.50 & 91.80 & 96.10 & 96.20 & 90.80 & - \\ \hline
\textbf{Proposed} & \textbf{94.40} & \textbf{95.30} & \textbf{96.90} & \textbf{97.10} & \textbf{93.40} & \textbf{96.07} \\ \hline
\end{tabular}

\end{table}
\endgroup

\section{Conclusion and Future Research Directions}
This research presents a deep learning based video kinship verification framework using a novel Supervised Mixed Norm Autoencoder.  
The proposed three-stage SMNAE encodes the spatio-temporal kinship cues present in video frames for video-based kinship verification. The comparative analysis with existing algorithms on publicly available video-based and five image-based kinship databases demonstrates the efficacy of the proposed kinship verification framework. Also, a new kinship video database, termed as KIVI database, is introduced which contains video sequences of 211 families, 355 true kin pairs, and seven kin-relations with more than 250,000 still frame images. On the KIVI database, the proposed algorithm yields state-of-the-art kinship verification accuracy of 83.18\%. In the future, this kinship database will be expanded to include new kin-relations such as cousins, nieces, nephews, aunts, and uncles and further evaluation will be performed on Family in the Wild database. Finally, we also plan to extend the proposed algorithm by incorporating video frame selection techniques \cite{mdlface,featureframe}. 


\section*{Appendix}

Additional experiments are performed to evaluate the efficacy of the proposed SMNAE on two standard databases, MNIST and CIFAR-10. Since the input is a single image, a minor modification is required. In Eq. 4, the parameter, $M_{ij}$ is updated to denote class information and rest of the computation remains the same. The experiments are performed with the pre-defined experimental protocol of the MNIST and CIFAR-10 databases, and the performance of the proposed SMNAE is compared with the traditional KL-divergence based Autoencoder (KLD) and Group Sparse Autoencoder (GSAE) [48]. As shown in Table \ref{tab:appresults}, on MNIST database, SMNAE yields the lowest error of 0.98\% as compared to 1.19\% by GSAE, and 1.71\% by KL-divergence based autoencoder. On CIFAR-10 image database, SMNAE achieves the highest classification accuracy of 79.50\% which is at least 2.7\% higher than other autoencoder formulations. These experiments further show the usefulness of $\ell_{2,p}$-norm based SMNAE approach on other object classification problems.


\begingroup
\renewcommand{\arraystretch}{1}    
\begin{table}[!t]
\centering
\caption{Performance of the proposed SMNAE, KLD, and GSAE
autoencoders on MNIST and CIFAR-10 databases.}
\label{tab:appresults}
\begin{tabular}{|l|l|c|c|c|}
\hline
\textbf{Database} & \textbf{Metric}          & \textbf{KLD}   & \textbf{GSAE}  &  \textbf{SMNAE} \\ \hline
MNIST    & Error rate (\%) & 1.71  & 1.19  & \textbf{0.98}           \\ \hline
CIFAR-10 & Accuracy (\%)   & 74.30 & 76.80 & \textbf{79.50}       \\ \hline 
\end{tabular}
\end{table}
\endgroup

\section*{Acknowledgement}
The authors acknowledge the support of NVIDIA for the donation of the Tesla K40 and Titan X Pascal GPUs.

\bibliographystyle{IEEEtran}
\bibliography{egbib}

\begin{thebibliography}{10}
\providecommand{\url}[1]{#1}
\csname url@samestyle\endcsname
\providecommand{\newblock}{\relax}
\providecommand{\bibinfo}[2]{#2}
\providecommand{\BIBentrySTDinterwordspacing}{\spaceskip=0pt\relax}
\providecommand{\BIBentryALTinterwordstretchfactor}{4}
\providecommand{\BIBentryALTinterwordspacing}{\spaceskip=\fontdimen2\font plus
\BIBentryALTinterwordstretchfactor\fontdimen3\font minus
  \fontdimen4\font\relax}
\providecommand{\BIBforeignlanguage}[2]{{%
\expandafter\ifx\csname l@#1\endcsname\relax
\typeout{** WARNING: IEEEtran.bst: No hyphenation pattern has been}%
\typeout{** loaded for the language `#1'. Using the pattern for}%
\typeout{** the default language instead.}%
\else
\language=\csname l@#1\endcsname
\fi
#2}}
\providecommand{\BIBdecl}{\relax}
\BIBdecl

\bibitem{Fang2010}
R.~Fang, K.~D. Tang, N.~Snavely, and T.~Chen, ``Towards computational models of
  kinship verification,'' in \emph{IEEE International Conference on Image
  Processing}, 2010, pp. 1577--1580.

\bibitem{Siyu2011}
S.~Xia, M.~Shao, and Y.~Fu, ``Kinship verification through transfer learning,''
  in \emph{International Joint Conference on Artificial Intelligence}, 2011,
  pp. 2539--2544.

\bibitem{Zhou2011}
X.~Zhou, J.~Hu, J.~Lu, Y.~Shang, and Y.~Guan, ``Kinship verification from
  facial images under uncontrolled conditions,'' in \emph{ACM Multimedia},
  2011, pp. 953--956.

\bibitem{Ming_CVPR11_Genealogical}
M.~Shao, S.~Xia, and Y.~Fu, ``Genealogical face recognition based on {UB}
  {K}in{F}ace database,'' in \emph{IEEE Computer Vision and Pattern Recognition
  Workshops}, 2011, pp. 60--65.

\bibitem{Zhou2012}
X.~Zhou, J.~Lu, J.~Hu, and Y.~Shang, ``Gabor-based gradient orientation pyramid
  for kinship verification under uncontrolled environments,'' in \emph{ACM
  International Conference on Multimedia}, 2012, pp. 725--728.

\bibitem{xia2012understanding}
S.~Xia, M.~Shao, J.~Luo, and Y.~Fu, ``Understanding kin relationships in a
  photo,'' \emph{IEEE Transactions on Multimedia}, vol.~14, no.~4, pp.
  1046--1056, 2012.

\bibitem{Kohli2012}
N.~Kohli, R.~Singh, and M.~Vatsa, ``Self-similarity representation of {W}eber
  faces for kinship classification,'' in \emph{IEEE Conference on Biometrics:
  Theory, Applications and Systems}, 2012, pp. 245--250.

\bibitem{fang2013kinship}
R.~Fang, A.~C. Gallagher, A.~Loui, and T.~Chen, ``Kinship classification by
  modeling facial feature heredity,'' in \emph{International Conference on
  Image Processing}, 2013, pp. 2983--2987.

\bibitem{lu}
J.~Lu, X.~Zhou, Y.-P. Tan, Y.~Shang, and J.~Zhou, ``Neighborhood repulsed
  metric learning for kinship verification,'' \emph{IEEE Transactions on
  Pattern Analysis and Machine Intelligence}, vol.~36, no.~2, pp. 331--345,
  2014.

\bibitem{hu2014large}
J.~Hu, J.~Lu, J.~Yuan, and Y.-P. Tan, ``Large margin multi-metric learning for
  face and kinship verification in the wild,'' in \emph{Asian Conference on
  Computer Vision}.\hskip 1em plus 0.5em minus 0.4em\relax Springer, 2014, pp.
  252--267.

\bibitem{Yan}
H.~Yan, J.~Lu, W.~Deng, and X.~Zhou, ``Discriminative multimetric learning for
  kinship verification,'' \emph{IEEE Transactions on Information Forensics and
  Security}, vol.~9, no.~7, pp. 1169--1178, 2014.

\bibitem{dehghanOVS14}
A.~Dehghan, E.~G. Ortiz, R.~Villegas, and M.~Shah, ``{W}ho {D}o {I} {L}ook
  {L}ike? {D}etermining parent-offspring resemblance via gated autoencoders,''
  in \emph{IEEE Computer Vision and Pattern Recognition}, 2014, pp. 1757--1764.

\bibitem{guo2014graph}
Y.~Guo, H.~Dibeklioglu, and L.~van~der Maaten, ``Graph-based kinship
  recognition,'' in \emph{IEEE International Conference on Pattern
  Recognition}, 2014, pp. 4287--4292.

\bibitem{Yan2014}
H.~Yan, J.~Lu, and X.~Zhou, ``Prototype-based discriminative feature learning
  for kinship verification,'' \emph{IEEE Transactions on {C}ybernetics},
  vol.~45, no.~11, pp. 2535--2545, 2015.

\bibitem{fisher_kinship}
Q.~Liu, A.~Puthenputhussery, and C.~Liu, ``Inheritable {F}isher vector feature
  for kinship verification,'' in \emph{IEEE International Conference on
  Biometrics Theory, Applications and Systems}, 2015, pp. 1--6.

\bibitem{genetic_kinship}
P.~Alirezazadeh, A.~Fathi, and F.~Abdali-Mohammadi, ``A genetic algorithm-based
  feature selection for kinship verification,'' \emph{IEEE Signal Processing
  Letters}, vol.~22, no.~12, pp. 2459--2463, 2015.

\bibitem{qin}
X.~Qin, X.~Tan, and S.~Chen, ``Tri-subject kinship verification: Understanding
  the core of a family,'' \emph{IEEE Transactions on Multimedia}, vol.~17,
  no.~10, pp. 1855--1867, 2015.

\bibitem{Zhou2015}
X.~Zhou, Y.~Shang, H.~Yan, and G.~Guo, ``Ensemble similarity learning for
  kinship verification from facial images in the wild,'' \emph{Information
  Fusion}, vol.~32, pp. 40--48, 2016.

\bibitem{robinson2016families}
J.~P. Robinson, M.~Shao, Y.~Wu, and Y.~Fu, ``Families in the wild ({FIW}):
  Large-scale kinship image database and benchmarks,'' in \emph{ACM Conference
  on Multimedia}, 2016, pp. 242--246.

\bibitem{xuandshang}
M.~Xu and Y.~Shang, ``Kinship measurement on face images by structured
  similarity fusion,'' \emph{IEEE Access}, vol.~4, pp. 10\,280--10\,287, 2016.

\bibitem{wuusefulness}
X.~Wu, E.~Boutellaa, M.~B. L{\'o}pez, X.~Feng, and A.~Hadid, ``On the
  usefulness of color for kinship verification from face images,'' in
  \emph{IEEE International Workshop on Information Forensics and Security},
  2016, pp. 1--6.

\bibitem{Yan2016}
H.~Yan, ``Kinship verification using neighborhood repulsed correlation metric
  learning,'' \emph{Image and Vision Computing}, vol.~60, pp. 91--97, 2017.

\bibitem{comments2015}
M.~B. Lopez, E.~Boutellaa, and A.~Hadid, ``Comments on the "{K}inship {F}ace in
  the {W}ild" data sets,'' \emph{IEEE Transactions on Pattern Analysis and
  Machine Intelligence}, vol.~38, no.~11, pp. 2342--2344, 2016.

\bibitem{LiFWXH16}
L.~Li, X.~Feng, X.~Wu, Z.~Xia, and A.~Hadid, ``Kinship verification from faces
  via similarity metric based convolutional neural network,'' in
  \emph{International Conference on Image Analysis and Recognition}, 2016, pp.
  539--548.

\bibitem{wang2017kinship}
S.~Wang, J.~P. Robinson, and Y.~Fu, ``Kinship verification on families in the
  wild with marginalized denoising metric learning,'' in \emph{IEEE
  International Conference on Automatic Face \& Gesture Recognition}, 2017, pp.
  216--221.

\bibitem{lu2017discriminative}
J.~Lu, J.~Hu, and Y.-P. Tan, ``Discriminative deep metric learning for face and
  kinship verification,'' \emph{IEEE Transactions on Image Processing},
  vol.~26, no.~9, pp. 4269--4282, 2017.

\bibitem{liu2017}
H.~Liu, J.~Cheng, and F.~Wang, ``Kinship verification based on status-aware
  projection learning,'' in \emph{IEEE International Conference on Image
  Processing}, 2017, pp. 1072--1076.

\bibitem{kohli16}
N.~Kohli, M.~Vatsa, R.~Singh, A.~Noore, and A.~Majumdar, ``Hierarchical
  representation learning for kinship verification,'' \emph{IEEE Transactions
  on Image Processing}, vol.~26, no.~1, pp. 289--302, 2017.

\bibitem{mahpod2018}
S.~Mahpod and Y.~Keller, ``Kinship verification using multiview hybrid distance
  learning,'' \emph{Computer Vision and Image Understanding}, vol. 167, pp. 28
  -- 36, 2018.

\bibitem{uvanemo}
H.~Dibeklioglu, A.~A. Salah, and T.~Gevers, ``Like father, like son: Facial
  expression dynamics for kinship verification,'' in \emph{IEEE International
  Conference on Computer Vision}, 2013, pp. 1497--1504.

\bibitem{dibeklioglu2017visual}
H.~Dibeklioglu, ``Visual transformation aided contrastive learning for
  video-based kinship verification,'' in \emph{IEEE Conference on International
  Conference on Computer Vision}, 2017, pp. 2459--2468.

\bibitem{dna}
A.~Kofman, ``{The Troubling Rise of Rapid DNA Testing},'' \url{goo.gl/UuhZSN},
  2016, [Online; accessed November 14, 2016].

\bibitem{stat_Youtube}
K.~Wagner, ``Facebook says video is huge -- 100-million-hours-per-day huge,''
  \url{goo.gl/LvZb51}, 2017, [Online; accessed March 14, 2017].

\bibitem{pfister2011}
T.~Pfister, X.~Li, G.~Zhao, and M.~Pietikäinen, ``Differentiating spontaneous
  from posed facial expressions within a generic facial expression recognition
  framework,'' in \emph{IEEE International Conference on Computer Vision
  Workshops}, 2011, pp. 868--875.

\bibitem{ng2011sparse}
A.~Ng, ``Sparse autoencoder,'' \emph{CS294A Lecture notes}, vol.~72, no. 2011,
  pp. 1--19, 2011.

\bibitem{gao2015single}
S.~Gao, Y.~Zhang, K.~Jia, J.~Lu, and Y.~Zhang, ``Single sample face recognition
  via learning deep supervised autoencoders,'' \emph{IEEE Transactions on
  Information Forensics and Security}, vol.~10, no.~10, pp. 2108--2118, 2015.

\bibitem{majumdar_pami_2016}
A.~Majumdar, R.~Singh, and M.~Vatsa, ``Face verification via class sparsity
  based supervised encoding,'' \emph{IEEE Transactions on Pattern Analysis and
  Machine Intelligence}, vol.~39, no.~6, pp. 1273--1280, 2017.

\bibitem{chartrand2007exact}
R.~Chartrand, ``Exact reconstruction of sparse signals via nonconvex
  minimization,'' \emph{IEEE Signal Processing Letters}, vol.~14, no.~10, pp.
  707--710, 2007.

\bibitem{chartrand2008iteratively}
R.~Chartrand and W.~Yin, ``Iteratively reweighted algorithms for compressive
  sensing,'' in \emph{IEEE International Conference on Acoustics, Speech and
  Signal Processing}, 2008, pp. 3869--3872.

\bibitem{yanl2p}
Y.~Yan, Z.~Xu, G.~Liu, Z.~Ma, and N.~Sebe, ``Glocal structural feature
  selection with sparsity for multimedia data understanding,'' in \emph{ACM
  International Conference on Multimedia}, 2013, pp. 537--540.

\bibitem{jenatton2010proximal}
R.~Jenatton, J.~Mairal, F.~R. Bach, and G.~R. Obozinski, ``Proximal methods for
  sparse hierarchical dictionary learning,'' in \emph{International Conference
  on Machine Learning}, 2010, pp. 487--494.

\bibitem{nesterov2007gradient}
Y.~Nesterov, ``Gradient methods for minimizing composite objective function,''
  UCL, Tech. Rep., 2007.

\bibitem{Zhang2014}
M.~Zhang, C.~Ding, Y.~Zhang, and F.~Nie, ``Feature selection at the discrete
  limit,'' in \emph{AAAI Conference on Artificial Intelligence}, 2014, pp.
  1355--1361.

\bibitem{cortes1995support}
C.~Cortes and V.~Vapnik, ``Support-vector networks,'' \emph{Machine learning},
  vol.~20, no.~3, pp. 273--297, 1995.

\bibitem{boutellaa_2016}
E.~Boutellaa, M.~B. L{\'{o}}pez, S.~Ait{-}Aoudia, X.~Feng, and A.~Hadid,
  ``Kinship verification from videos using spatio-temporal texture features and
  deep learning,'' \emph{CoRR}, vol. abs/1708.04069, 2017.

\bibitem{zhao2007dynamic}
G.~Zhao and M.~Pietikainen, ``Dynamic texture recognition using local binary
  patterns with an application to facial expressions,'' \emph{IEEE Transactions
  on Pattern Analysis and Machine Intelligence}, vol.~29, no.~6, 2007.

\bibitem{parkhi2015deep}
O.~Parkhi, V.~Andrea, and Z.~Andrew, ``Deep face recognition,'' in
  \emph{Proceedings of the British Machine Vision Conference}, 2015, pp.
  41.1--41.12.

\bibitem{kaminski2009human}
G.~Kaminski, S.~Dridi, C.~Graff, and E.~Gentaz, ``Human ability to detect
  kinship in strangers' faces: effects of the degree of relatedness,''
  \emph{Proceedings of the Royal Society of London B: Biological Sciences},
  vol. 276, no. 1670, pp. 3193--3200, 2009.

\bibitem{mdlface}
G.~Goswami, R.~Bhardwaj, R.~Singh, and M.~Vatsa, ``Mdlface: Memorability
  augmented deep learning for video face recognition,'' in \emph{IEEE
  International Joint Conference on Biometrics}, 2014, pp. 1--7.

\bibitem{featureframe}
G.~Goswami, M.~Vatsa, and R.~Singh, ``Face verification via learned
  representation on feature-rich video frames,'' \emph{IEEE Transactions on
  Information Forensics and Security}, vol.~12, no.~7, pp. 1686--1698, 2017.

\end{thebibliography}

\begin{IEEEbiography}[{\includegraphics[width=1in,height=1.25in,clip,keepaspectratio]{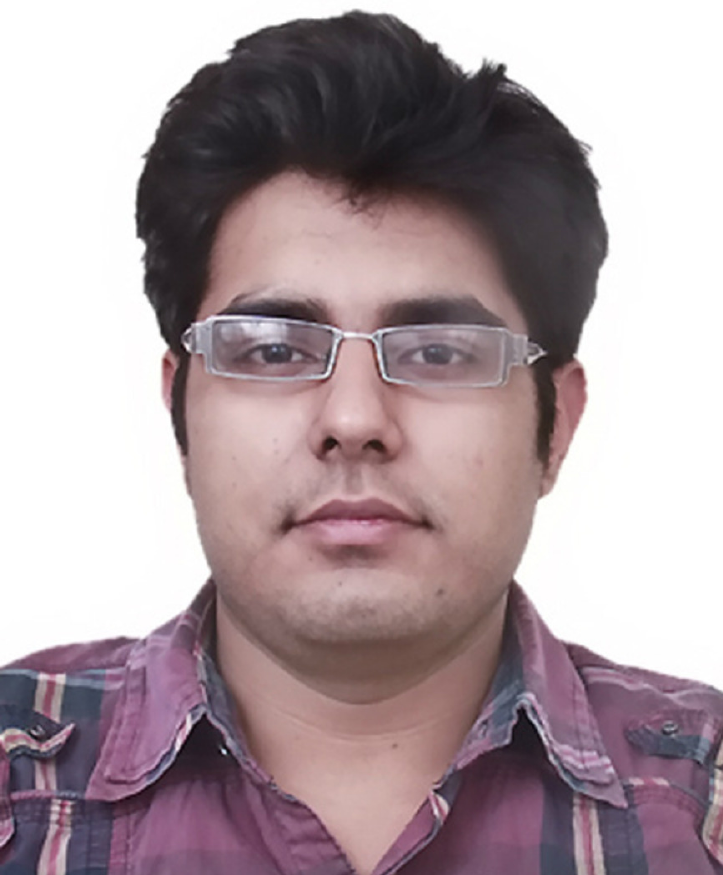}}]{Naman Kohli}
	received the B.Tech. (Hons.) degree in computer science from the IIIT-Delhi, India, in 2013. He is currently pursuing the Ph.D. degree in computer science with the Lane Department of Computer Science and Electrical Engineering, West Virginia University, USA. His research interests include biometrics, computer vision, and deep learning. He is a member of the Phi Kappa Phi, Upsilon Pi Epsilon, and the Eta Kappa Nu Honor Societies. He is a recipient of the Best Paper award at WACV 2016 and Best Poster award at IJCB 2017. 

\end{IEEEbiography}

\begin{IEEEbiography}[{\includegraphics[width=1in,height=1.25in,clip,keepaspectratio]{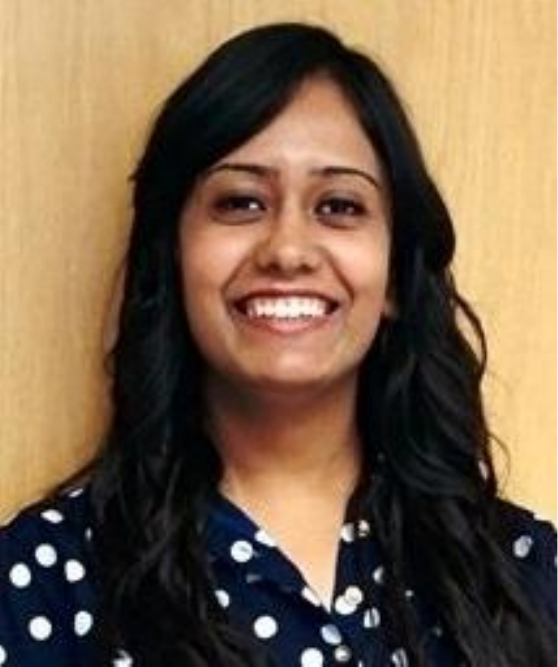}}]{Daksha Yadav}
received the B. Tech (Hons.) in Computer Science in 2013 from
IIIT-Delhi, India. She is currently pursuing the Ph.D. degree from the Lane Department of Computer
Science and Electrical Engineering at West Virginia University, USA. Her research interests include machine learning, biometrics, computer vision, and neurocognition. She is the recipient of multiple best paper and poster awards. She is also a member of the Phi Kappa Phi, Eta Kappa Nu, and Upsilon Pi Epsilon societies. 

\end{IEEEbiography}

\begin{IEEEbiography}[{\includegraphics[width=1in,height=1.25in,clip,keepaspectratio]{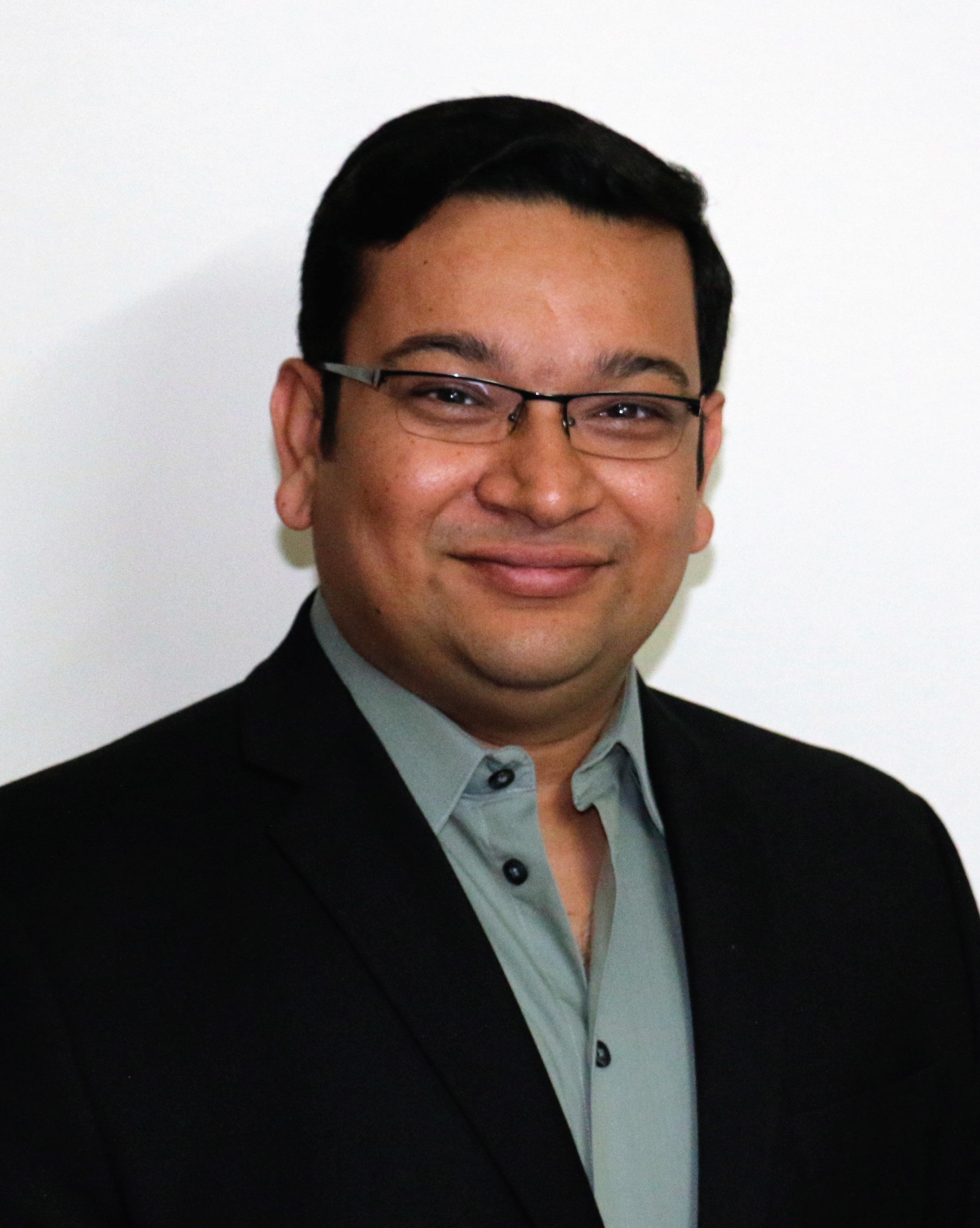}}]{Mayank Vatsa}
received the M.S. and Ph.D. degrees in computer science from West Virginia University, USA, in 2005 and 2008, respectively. He is currently Head of Infosys Center for Artificial Intelligence, Associate Professor with IIIT-Delhi, India, and Adjunct Associate Professor at West Virginia University, USA. He was a recipient of the FAST Award Project by DST (India) and several Best Paper and Best Poster Awards at international conferences. He has co-edited a book titled Deep learning in Biometrics and co-authored over 225 research papers. His areas of interest are biometrics, image processing, machine learning, and information fusion. He is the Vice President (Publications) of IEEE Biometrics Council where he started IEEE Transactions on Biometrics, Behavior, and Identity Science (IEEE T-BIOM). He is a senior member of IEEE and the Association for Computing Machinery. He is also an Area Chair of the Information Fusion (Elsevier) journal, Associate Editor of IEEE Access, and the PC Co-Chair of the 2013 ICB and 2014 IJCB.
\end{IEEEbiography}

\begin{IEEEbiography}[{\includegraphics[width=1in,height=1.25in,clip,keepaspectratio]{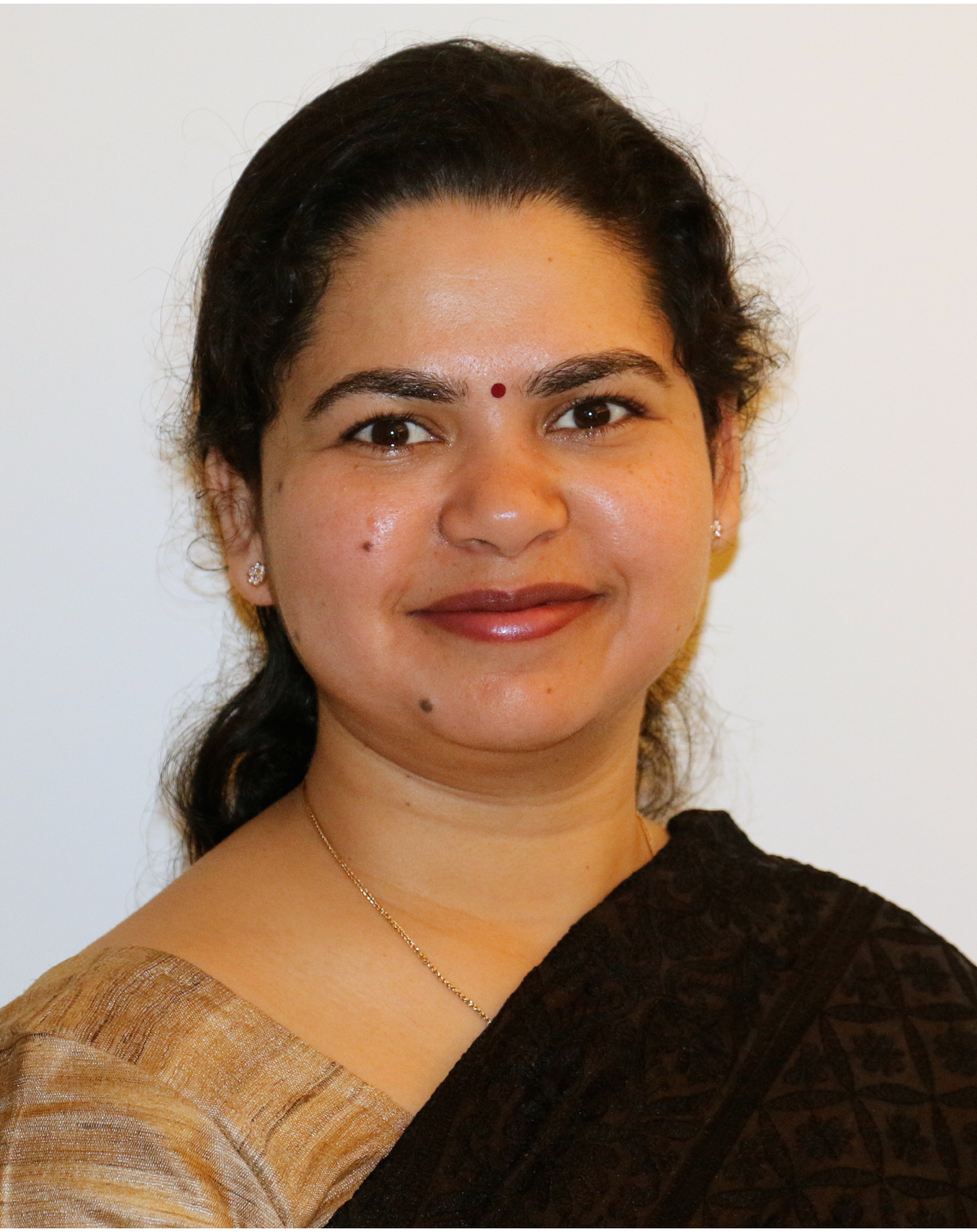}}]{Richa Singh}
received a Ph.D. degree in Computer Science from West Virginia University, Morgantown, USA in 2008. She is currently Associate Dean of Alumni and Communications, Associate Professor with IIIT-Delhi, India, and Adjunct Associate Professor at West Virginia University. She is a Fellow of IAPR and Senior Member of IEEE and ACM. Her areas of interest are pattern recognition, machine learning, and biometrics. She is a recipient of the Kusum and Mohandas Pai Faculty Research Fellowship at the IIIT-Delhi, the FAST Award by the Department of Science and Technology, India, and several best paper and best poster awards in international conferences. She has co-edited book titled Deep Learning in Biometrics and has delivered  tutorials on deep learning and domain adaptation in ICCV 2017, AFGR 2017, and IJCNN 2017. She is  an Editorial Board Member of Information Fusion (Elsevier), Associate Editor of Pattern Recognition, CVIU, IEEE Access, and the EURASIP Journal on Image and Video Processing (Springer). She has also served as the Program Co-Chair of BTAS 2016, IWBF 2018, and General Co-Chair of ISBA 2017. She is currently serving as Program Co-Chair of AFGR 2019.

\end{IEEEbiography}

\begin{IEEEbiography}[{\includegraphics[width=1in,height=1.25in,clip,keepaspectratio]{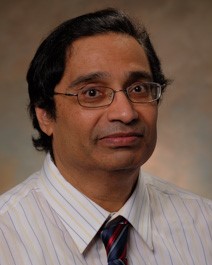}}]{Afzel Noore} received the Ph.D. degree in electrical engineering from West Virginia University. He was a Digital Design Engineer with Philips, India. He held several administrative positions as Associate Dean for Academic Affairs, Special Assistant to the Dean in the Statler College of Engineering and Mineral Resources, and as Associate Chair of the Lane Department of Computer Science and Electrical Engineering at West Virginia University. He currently serves as Professor and Associate Dean for Undergraduate Affairs in the Frank H. Dotterweich College of Engineering at Texas A\&M University-Kingsville. His research interests include computational intelligence, biometrics, machine learning, and pattern recognition. His research has been funded by NASA, the National Science Foundation, Westinghouse, General Electric, the Electric Power Research Institute, the U.S. Department of Energy, the U.S. Department of Justice, the Department of Defense, Army Research Laboratory, and the FBI. He has around 140 publications in refereed journals, book chapters, and conferences. He is a member of Phi Kappa Phi Honor Society, Sigma Xi Honor Society, Eta Kappa Nu Honor Society, and Tau Beta Pi Honor Society. He is a recipient of several teaching awards and research awards at West Virginia University. He has received 10 best paper and best poster awards in international conferences. He serves on the Editorial Board of the International Journal of Multimedia Intelligence and Security.

\end{IEEEbiography}

\end{document}